\documentclass[]{interact}

\usepackage{url}
\usepackage{subfigure}
\usepackage{epstopdf}% To incorporate .eps illustrations using PDFLaTeX, etc.
\usepackage[caption=false]{subfig}% Support for small, `sub' figures and tables

\usepackage[table, xcdraw]{xcolor}% Alternating row colours in tables
\usepackage{natbib}% Citation support using natbib.sty
\bibpunct[,
]{(}{)}{;}{a}{}{,}% Citation support using natbib.sty
\renewcommand{\bibfont}{\fontsize{10}{12}\selectfont}% Bibliography support using natbib.sty

\theoremstyle{plain}% Theorem-like structures provided by amsthm.sty

\theoremstyle{definition}

\theoremstyle{remark}

\begin{document}
    \makeatletter \gdef\@title{Low-Cost Stereo Vision for Robust 3D Positioning of Thin Radiata Pine Branches in Autonomous Drone Pruning}

    \author{\name{Yida Lin\textsuperscript{a}, Bing Xue\textsuperscript{a}, Mengjie Zhang\textsuperscript{a}, Sam Schofield\textsuperscript{b}, Richard Green\textsuperscript{b}}
    \affil{\textsuperscript{a} Centre of Data Science and Artificial Intelligence $\&$ School of Engineering and Computer Science, Victoria University of Wellington, PO Box 600, Wellington, New Zealand}
    \affil{\textsuperscript{b} Computer Vision Research Lab $\&$ Department of Computer Science and Software Engineering, Canterbury University, PO Box 8041, Christchurch, New Zealand}}

    \maketitle
    \begin{abstract}
        Manual pruning of radiata pine, a species of major economic importance
        to New Zealand forestry, is hazardous, labour-intensive, and increasingly
        constrained by workforce shortages. Existing autonomous pruning
        platforms typically rely on expensive sensors such as LiDAR and are
        limited to thick branches, which restricts their wider adoption. This
        paper investigates whether a single low-cost stereo camera mounted on a
        drone can provide sufficiently accurate branch detection and three-dimensional
        positioning to support autonomous pruning of branches as thin as 10\,mm,
        thereby removing the need for auxiliary depth sensors. The proposed
        pipeline comprises two stages: branch segmentation and depth estimation.
        For segmentation, Mask~R-CNN variants and the YOLOv8 and YOLOv9 families
        are compared on a custom dataset of 71 stereo image pairs captured with a
        ZED Mini camera; YOLOv8 and YOLOv9 are selected as representative state-of-the-art
        real-time segmentors at the time of data collection, and the framework is
        designed to remain compatible with newer YOLO releases. For depth
        estimation, a traditional method (SGBM with WLS filtering) and deep-learning-based
        methods (PSMNet, ACVNet, GWCNet, MobileStereoNet, RAFT-Stereo, and NeRF-Supervised
        Deep Stereo) are evaluated, including cross-dataset fine-tuning experiments
        that expose the domain gap between urban driving benchmarks and natural forestry
        scenes. The main novelty of this work lies in coupling stereo
        segmentation with a centroid-based triangulation algorithm and Median-Absolute-Deviation
        outlier rejection that converts a segmentation mask and disparity map
        into a single robust branch-to-camera distance, addressing the challenges
        of sparse texture, thin structures, and noisy disparity values typical of
        forest scenes. Qualitative evaluations at distances of 1--2\,m show that
        the learning-based stereo methods produce more coherent depth estimates than
        SGBM, demonstrating the feasibility of low-cost stereo vision for
        automated branch positioning in forestry.
    \end{abstract}


    \section{Introduction}

    Pinus radiata, commonly known as radiata pine, is a highly valuable species
    extensively cultivated in New Zealand owing to its rapid growth and broad
    applicability across forestry and timber industries. The species supplies high-quality
    timber for construction, paper manufacturing, and other wood-based products,
    and contributes substantially to regional economies \citep{lin2024drone,mason2023impacts}.
    As an indication of scale, the radiata pine industry in New South Wales,
    Australia, contributed approximately \$3 billion in 2021--22\footnote{https://www.dpi.nsw.gov.au/dpi/climate/climate-vulnerability-assessment/forestry/radiata-pine}.
    To produce strong, straight trunks with clear, knot-free wood, regular
    pruning of lower branches is required. Manual pruning, however, remains one of
    the most hazardous occupations in forestry: in the United States, the Bureau
    of Labor Statistics records a fatality rate of approximately 110 per 100{,}000
    tree trimmers and pruners\footnote{https://tcimag.tcia.org/safety/tree-work-safety-by-the-numbers/},
    roughly 30 times the all-industry average, while non-fatal injury rates
    reach around 239 per 10{,}000 workers, compared with 89 per 10{,}000 across all
    industries. These risks, combined with persistent labour shortages in the sector,
    have created a strong demand for safer, automated alternatives.

    Drones are now used for an increasing range of forestry tasks such as inventory,
    health monitoring, and spraying, yet autonomous in-canopy \emph{pruning} remains
    comparatively under-developed. Existing drone- and manipulator-based pruning
    prototypes typically combine LiDAR or multi-camera rigs with heavy-duty cutters,
    which raises hardware cost and constrains the achievable branch thickness. Early
    demonstrations such as \citet{molina2017aerial} target relatively coarse
    branches, and although sensing and cutting hardware have advanced considerably
    since, the published state of the art still concentrates on branches in the
    order of 20--40\,mm or thicker rather than on the fine, high-canopy branches
    that matter most for clear-wood production. Reducing the minimum detectable branch
    diameter, rather than building yet another cutting platform, is therefore the
    perception bottleneck addressed in this paper.

    Motivated by this thickness gap, and building on our IVCNZ 2024 conference
    paper \citep{lin2024drone}, this study aims to detect and three-dimensionally
    \emph{position} radiata pine branches as thin as 10\,mm at working distances
    of 1--2\,m, using only a single low-cost stereo camera and no auxiliary depth
    sensors. The mechanical cutting itself is performed by a robotic arm
    developed by partner institutions and is outside the scope of this paper; here
    we focus exclusively on the perception pipeline that locates each branch in three-dimensional
    space and supplies the cutting tool with an accurate target position.
    Consistent with the title, the central task is branch \emph{positioning} (i.e.~recovering
    the 3D location of a branch), of which segmentation-based detection is one
    component; obtaining an accurate per-branch distance from a noisy stereo disparity
    map is the more challenging step and is the main contribution of this work.
    An overview of the wider drone platform developed by the project consortium
    is available at \url{https://ucvision.org.nz/drones/}.

    \paragraph{Relation to the IVCNZ 2024 conference paper.}
    An earlier version of this work appeared at IVCNZ 2024 \citep{lin2024drone},
    which reported preliminary segmentation results with a single YOLO variant and
    a single SGBM-based depth pipeline on a small set of indoor stereo images.
    The present paper substantially extends that work in three directions:
    \begin{enumerate}
        \item \textbf{Broader segmentation study.} The comparison is extended
            from a single YOLO variant to a side-by-side evaluation of seven
            Mask~R-CNN backbones and the full YOLOv8 and YOLOv9 segmentation
            families on an enlarged dataset of 71 ZED~Mini stereo pairs. YOLOv8 and
            YOLOv9 are chosen as representative state-of-the-art real-time segmentors
            at the time of data collection; the framework remains directly
            compatible with newer YOLO releases as they appear.

        \item \textbf{Comprehensive stereo depth analysis.} The depth-estimation
            analysis is extended from SGBM alone to six deep-learning stereo
            networks (PSMNet, ACVNet, GWCNet, MobileStereoNet, RAFT-Stereo, and
            NeRF-Supervised Deep Stereo), including cross-dataset fine-tuning
            experiments on KITTI~2012, KITTI~2015, and Scene~Flow that quantify
            the domain gap between urban driving benchmarks and natural forestry
            scenes.

        \item \textbf{New centroid-based triangulation with MAD filtering.} A
            new centroid-based triangulation procedure with Median-Absolute-Deviation
            outlier rejection is introduced to fuse the segmentation mask and
            disparity map into a single robust branch-to-camera distance. This component
            was not present in the conference version and is what enables reliable
            distance estimation on thin, sparsely textured branches.
    \end{enumerate}

    The primary objective of the perception pipeline is therefore to provide
    accurate and reliable 3D positions for radiata pine branches as thin as 10\,mm
    at working distances of 1--2\,m, using only a passive stereo camera. The remainder
    of the paper is organised as follows. Section~\ref{sec:relatedwork} reviews
    related work on object segmentation and stereo depth estimation. Section~\ref{sec:methods}
    presents the proposed perception pipeline, including branch segmentation, traditional
    and deep-learning disparity estimation, and the centroid-based triangulation
    with MAD outlier rejection. Section~\ref{sec:data} describes the data collection
    protocol and its limitations. Section~\ref{sec:setup} reports the
    experimental setup, datasets, training configuration, and evaluation metrics.
    Section~\ref{sec:results} presents the results, discussion, and ablation
    analysis. Sections~\ref{sec:future} and~\ref{sec:conclusion} discuss future work
    and conclusions, respectively.

    \section{Related Work}
    \label{sec:relatedwork}

    %現在用機械對樹枝進行修剪的設備都非常貴重，而且其配備了非常多的設備例如多個高精度攝像頭，多個激光雷達，等等，這些機械雖然可以大大提升農業機械的工作效率，但是也讓農業機械的門檻變的非常高，在我們這個研究中，我們只用了一個立體攝像頭，後期可以用單個攝像頭。

    In contemporary agricultural machinery, pruning equipment is often associated
    with high costs and the integration of advanced technologies, such as
    multiple high-precision cameras and LiDAR sensors \citep{kulbacki2018survey,pakeerathan2023smart}.
    While these systems significantly enhance operational efficiency, they also contribute
    to increased costs and complexity in deployment, which may impede the
    commercial viability and broader adoption of drone-based solutions. To mitigate
    these challenges, this research emphasizes the use of a single stereo camera
    for branch detection and distance measurement between branches and the drone.

    Given the reliance on a single stereo camera, the study leverages advanced computer
    vision techniques to detect the branches, with a particular focus on two
    critical areas: branch detection\citep{zou2023object} and depth map generation\citep{laga2020survey}.
    For branch detection, the algorithms must be both highly efficient and precise,
    as the drone needs to identify branches in real-time during operation.

    To facilitate depth map generation, innovative methods are employed to accurately
    estimate the distance between the branches and the camera. By integrating
    branch detection with depth map generation, the spatial positioning of branches
    can be precisely determined. This accurate spatial information then enables
    the drone's robotic arm to execute branch trimming with precision, ensuring effective
    and targeted pruning.

    \subsection{Detection and Segmentation}

    Object detection \citep{zou2023object} and image segmentation \citep{minaee2021image}
    are two complementary tasks in computer vision. Detection localises objects with
    bounding boxes, whereas segmentation further delineates the pixel-level
    shape of each object. Because the downstream triangulation step in this work
    needs the precise pixel support of a branch (rather than only its bounding
    box), the perception pipeline is based on instance \emph{segmentation} rather
    than detection alone \citep{sharma2022survey}.

    Modern segmentors fall broadly into two families: two-stage region-based detectors
    and one-stage real-time detectors. The R-CNN lineage---R-CNN \citep{girshick2014rich},
    SPP-Net \citep{he2015spatial}, Fast R-CNN \citep{girshick2015fast}, Faster R-CNN
    with the Region Proposal Network \citep{ren2016faster}, and Mask R-CNN
    \citep{he2017mask}---progressively moved from external region proposals and per-region
    feature extraction to end-to-end region-proposal networks and a dedicated mask
    branch. Their main \emph{advantage} is high mask quality on small or thin objects
    thanks to RoIAlign and a separate segmentation head; the principal \emph{disadvantage}
    is inference cost and a non-trivial memory footprint, which makes them less
    attractive for on-board, real-time use on a drone. The YOLO family \citep{redmon2016you,reis2023real,wang2024yolov9}
    takes the opposite trade-off: a single-shot, fully convolutional architecture
    with anchor-free or decoupled heads delivers an order-of-magnitude faster inference
    at the cost of slightly weaker mask boundaries, especially for very thin
    structures.

    These contrasting properties motivate the model selection in this paper. Mask~R-CNN
    with seven backbones (R50/R101/X101 in C4, DC5, and FPN variants) is included
    as a strong, widely cited two-stage baseline that is known to produce high-quality
    masks. From the one-stage family, the YOLOv8 and YOLOv9 segmentation
    variants are selected as representative real-time segmentors that were state
    of the art at the time of data collection; we do not claim them to be the most
    recent YOLO release, and the framework is intentionally agnostic to the
    specific YOLO version, so that newer variants can be substituted in directly.
    Other one-stage segmentors (e.g.~SOLOv2, Mask2Former) were not evaluated
    here because the YOLO family already represents the speed end of the trade-off
    and Mask~R-CNN already represents the accuracy end; covering the full space exhaustively
    is left to future work.

    Segmentation quality in this paper is reported using the standard COCO mean Average
    Precision metric, $\mathrm{mAP}_{50\text{--}95}$ \citep{lin2014microsoft}.
    This is an \emph{existing} measure and is not redefined here; the precise
    computation, training protocol, and per-model results are deferred to
    Sections~\ref{sec:setup} and~\ref{sec:results}.

    %-------------------------------------------------------------------------
    \subsection{Depth Map}
    %這邊寫下在我們這個項目中depth map的作用
    Depth map generation \citep{laga2020survey} infers a scene's three-dimensional
    structure from one or more images and is a critical component of any autonomous
    pruning pipeline, because the robotic cutter must know not only \emph{which}
    pixels belong to a branch but also \emph{how far away} the branch is.

    Depth-sensing approaches are commonly grouped into \emph{active} and \emph{passive}
    methods. Active sensors emit a signal and measure its return, e.g.~LiDAR \citep{wu2018squeezeseg},
    structured light \citep{yang2024overview}, and time-of-flight cameras \citep{kolb2010time};
    they typically deliver high metric accuracy but increase payload, power
    consumption, and unit cost, and can struggle with thin, partially occluded structures
    such as small branches. Passive methods rely on existing optical information,
    including stereo matching \citep{wang2011obtaining}, multi-view geometry \citep{hartley2003multiple},
    and monocular depth estimation \citep{eigen2014depth}. Stereo matching offers
    a favourable trade-off for a low-cost drone platform: it provides metrically
    meaningful depth from a single passive sensor (a stereo rig such as the ZED~Mini)
    without active illumination, while avoiding the well-known scale ambiguity of
    monocular estimation. For this reason the system in this paper is built
    around binocular stereo matching, with the camera baseline and focal length fixed
    by the ZED~Mini hardware.

    \begin{figure}[htbp]
        \centering
        \includegraphics[width=12cm]{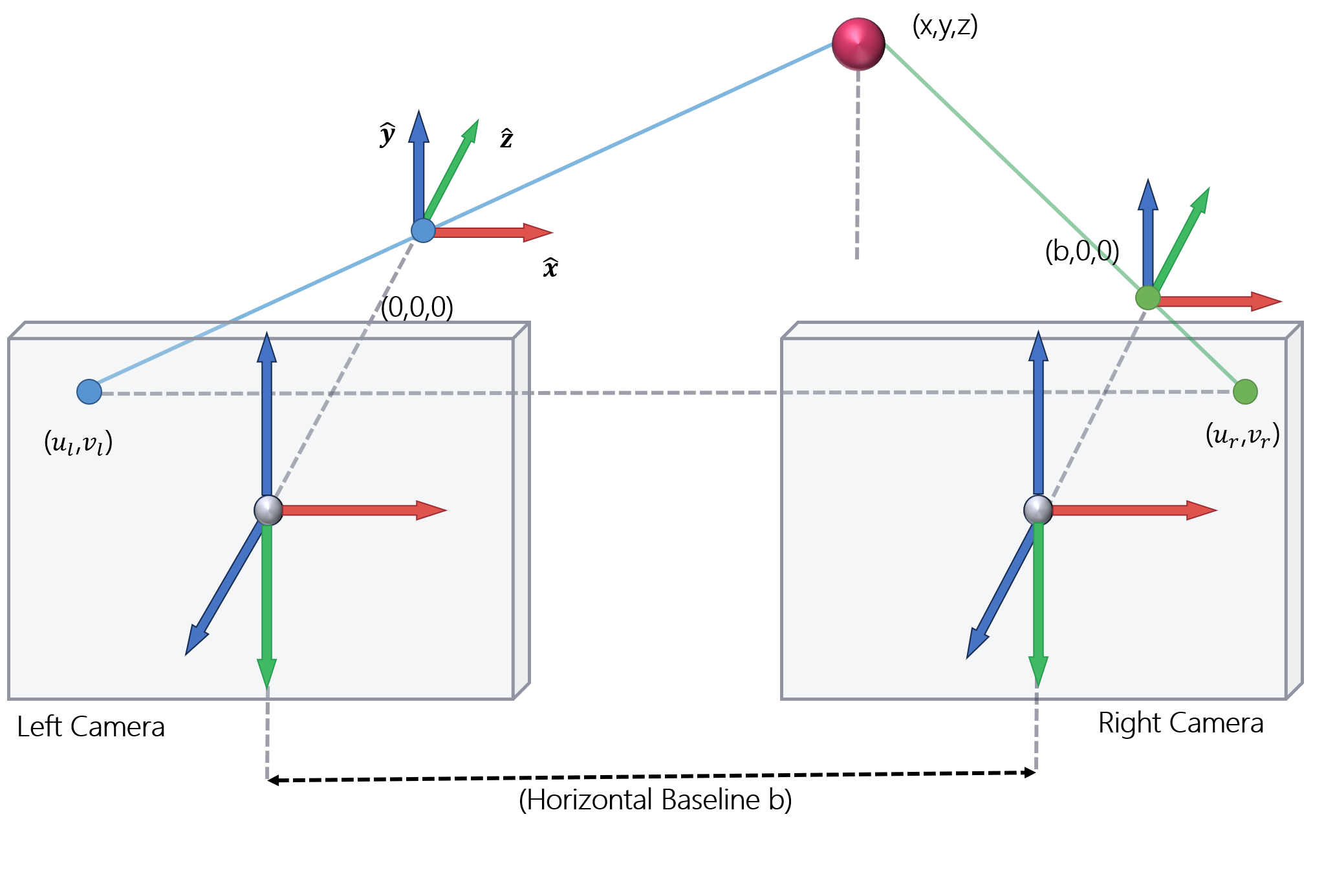}
        \caption{Triangulation using two cameras to obtain the depth map. The point
        $(u_{l}, v_{l})$ is the projection of point $p(x,y,z)$ onto the left image
        plane, and $(u_{r}, v_{r})$ is the projection of the same point onto the
        right image plane; $b$ is the baseline distance between the two cameras.
        Symbols are summarised in Table~\ref{tab:stereo_symbols}.}
        \label{fig:triangulation1}
    \end{figure}

    \begin{figure}[htbp]
        \centering
        \includegraphics[width=12cm]{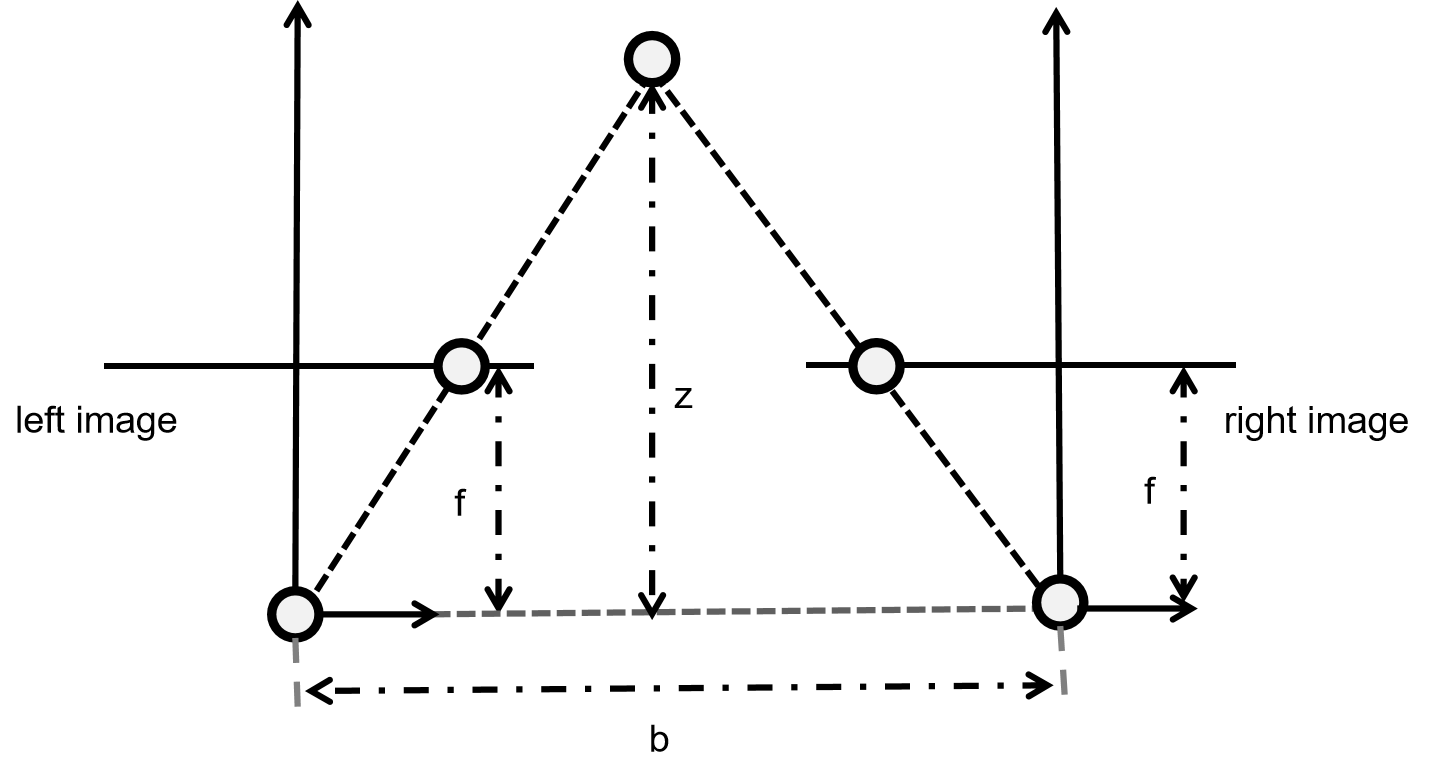}
        \caption{Triangulation geometry showing the focal length $f$. Given $f$,
        the baseline $b$, and the disparity of $p(x,y,z)$ between the left and right
        images, the depth $z$ of $p$ can be recovered. Symbols are summarised in
        Table~\ref{tab:stereo_symbols}.}
        \label{fig:triangulation2}
    \end{figure}

    \begin{table}[htbp]
        \centering
        \begin{tabular}{cl}
            \toprule \textbf{Symbol} & \textbf{Meaning}                                                                \\
            \midrule $f_{x}$         & Focal length of the camera in the $x$-direction.                                \\
            $f_{y}$                  & Focal length of the camera in the $y$-direction.                                \\
            $o_{x}$                  & Horizontal offset of the principal point from the top-left of the image sensor. \\
            $o_{y}$                  & Vertical offset of the principal point from the top-left of the image sensor.   \\
            $u_{l}, v_{l}$           & Horizontal/vertical coordinates of a scene point in the left image.             \\
            $u_{r}, v_{r}$           & Horizontal/vertical coordinates of the same scene point in the right image.     \\
            $b$                      & Baseline distance between the two camera centres.                               \\
            $d$                      & Disparity, $d = u_{l}- u_{r}$.                                                  \\
            $z$                      & Depth, i.e.~distance from the scene point to the camera.                        \\
            \bottomrule
        \end{tabular}
        \caption{Notation used in the stereo triangulation equations (Eqs.~\ref{eq:ul}--\ref{eq:depth_concise})
        and Figures~\ref{fig:triangulation1}--\ref{fig:triangulation2}.}
        \label{tab:stereo_symbols}
    \end{table}

    Equations~(\ref{eq:ul}) and (\ref{eq:vl}) describe the projection of a scene
    point onto the left image, and Eqs.~(\ref{eq:ur}) and (\ref{eq:vr}) onto the
    right image:

    \begin{equation}
        u_{l}= f_{x}\frac{x}{z}+o_{x}\label{eq:ul}
    \end{equation}

    \begin{equation}
        v_{l}= f_{y}\frac{y}{z}+o_{y}\label{eq:vl}
    \end{equation}

    \begin{equation}
        u_{r}= f_{x}\frac{x-b}{z}+o_{x}\label{eq:ur}
    \end{equation}

    \begin{equation}
        v_{r}= f_{y}\frac{y}{z}+o_{y}\label{eq:vr}
    \end{equation}

    \noindent
    Combining Eqs.~(\ref{eq:ul}) and (\ref{eq:vl}), as well as Eqs.~(\ref{eq:ur})
    and (\ref{eq:vr}), we obtain the pixel coordinates as Eq.~(\ref{eq:combined}).
    \begin{equation}
        \begin{aligned}
            (u_{l},v_{l}) = (f_{x}\frac{x}{z}+o_{x}, f_{y}\frac{y}{z}+o_{y})   \\
            (u_{r},v_{r}) = (f_{x}\frac{x-b}{z}+o_{x}, f_{y}\frac{y}{z}+o_{y})
        \end{aligned}
        \label{eq:combined}
    \end{equation}
    \noindent
    Based on the pixel coordinates of the left and right cameras, we can find the
    coordinates of the object in three dimensions $(x,y,z)$.
    \begin{equation}
        \begin{aligned}
            x = \frac{b(u_{l}-o_{x})}{(u_{l}-u_{r})}           \\
            y = \frac{bf_{x}(v_{l}-o_{y})}{f_{y}(u_{l}-u_{r})} \\
            z = \frac{bf_{x}}{(u_{l}-u_{r})}
        \end{aligned}
    \end{equation}

    \noindent
    From the above, the disparity $d$ and depth $z$ are defined as:

    \begin{equation}
        d = u_{l}- u_{r}
    \end{equation}

    \begin{equation}
        z = \frac{b f_{x}}{u_{l}-u_{r}}\label{eq:depth}
    \end{equation}

    \noindent
    We set the product of the baseline $b$ and the focal length of the camera in
    the x-direction $f_{x}$ to a new variable $W$.
    \begin{equation}
        W = b \cdot f_{x}
    \end{equation}
    Substituting $W$ into Eq.~(\ref{eq:depth}), we get the more concise Eq.~(\ref{eq:depth_concise}).
    \begin{equation}
        z = \frac{W}{d}\label{eq:depth_concise}
    \end{equation}
    Since $W$ is fixed, and $z$ and $d$ are inversely proportional, the larger the
    disparity value, the smaller the depth value. In other words, a larger
    disparity value indicates that the pixel point is closer to the camera.

    Among traditional stereo methods, Block Matching (BM) \citep{scharstein2002taxonomy}
    and Semi-Global Block Matching (SGBM) \citep{hirschmuller2007stereo} are the
    two dominant techniques. BM is a local search-based method that estimates
    disparity by finding the best match within a fixed window. Its main \emph{advantage}
    is real-time efficiency on commodity CPUs, but it is prone to errors in sparse-texture
    and occluded regions---precisely the conditions encountered around thin branches
    against a complex background. SGBM addresses these weaknesses through a semi-global
    cost-aggregation strategy that approximates a global energy minimisation
    along multiple scan-line directions, yielding smoother and more robust disparity
    maps in texture-rich scenes at moderate runtime cost. SGBM (with WLS post-filtering)
    is therefore retained in this work as the traditional baseline.

    Deep-learning-based stereo methods have substantially improved disparity
    accuracy at the cost of higher computational and data requirements. Cost-volume
    networks such as PSMNet \citep{chang2018pyramid} introduce pyramid pooling
    and 3D convolutions for richer context aggregation; GWCNet \citep{guo2019group}
    forms a group-wise correlation cost volume to reduce parameter count while
    preserving matching information; and ACVNet \citep{xu2022attention} adds an
    attention-based concatenation volume that further suppresses ambiguous
    matches. MobileStereoNet \citep{shamsafar2022mobilestereonet} re-engineers
    the cost-volume architecture for resource-constrained platforms, trading some
    accuracy for a much lower memory and compute footprint. RAFT-Stereo \citep{lipson2021raft}
    departs from a single-pass cost volume and instead iteratively refines disparity
    through gated recurrent updates, which improves accuracy on thin structures
    and large disparity ranges but increases per-frame latency. NeRF-Supervised Deep
    Stereo \citep{tosi2023nerf} sidesteps the cost of acquiring ground-truth
    disparity by training on stereo pairs rendered from Neural Radiance Fields, which
    is especially useful for application domains---such as forestry---where dense
    depth annotations are unavailable.

    These methods cover a representative range of stereo paradigms---cost-volume
    aggregation (PSMNet, GWCNet, ACVNet), lightweight efficient design (MobileStereoNet),
    iterative refinement with recurrent updates (RAFT-Stereo), and self-/NeRF-supervised
    training (NeRF-Supervised Deep Stereo)---and are therefore selected for evaluation
    on the radiata pine branch dataset. Quantitative metrics (e.g.~Root Mean
    Squared Error against reference disparity, where available) and the precise
    evaluation protocol are described in Section~\ref{sec:setup} and the results
    in Section~\ref{sec:results}.

    \section{Methods}
    \label{sec:methods}
    %我們把這個分成了3個部分。資料搜集，圖像分割，深度圖得到

    \subsection{Overview and Research Gap}
    \label{sec:overview}

    The reviewed literature shows two well-developed but largely independent
    research streams: instance segmentation, which yields high-quality 2D masks
    of objects, and stereo depth estimation, which yields a per-pixel disparity
    map. Neither stream, in isolation, returns the single quantity that an autonomous
    pruning end-effector actually needs: a robust \emph{branch-to-camera
    distance} for each detected branch. The challenge is non-trivial because (i)
    thin branches occupy only a few-pixel-wide region of the disparity map, where
    stereo matching is most error-prone; (ii) bark and foliage produce sparse, repetitive
    texture that violates the texture assumptions of classical stereo; and (iii)
    a small number of grossly wrong disparity pixels inside the branch mask can
    drift the mean distance by many centimetres, more than the tolerance of a robotic
    cutter. This paper addresses these issues with an integrated pipeline that
    explicitly couples segmentation and disparity through a centroid-based
    sampling and Median-Absolute-Deviation (MAD) outlier rejection step, which forms
    the main methodological contribution.

\begin{figure}[htbp]
        \centering
        \includegraphics[width=14cm]{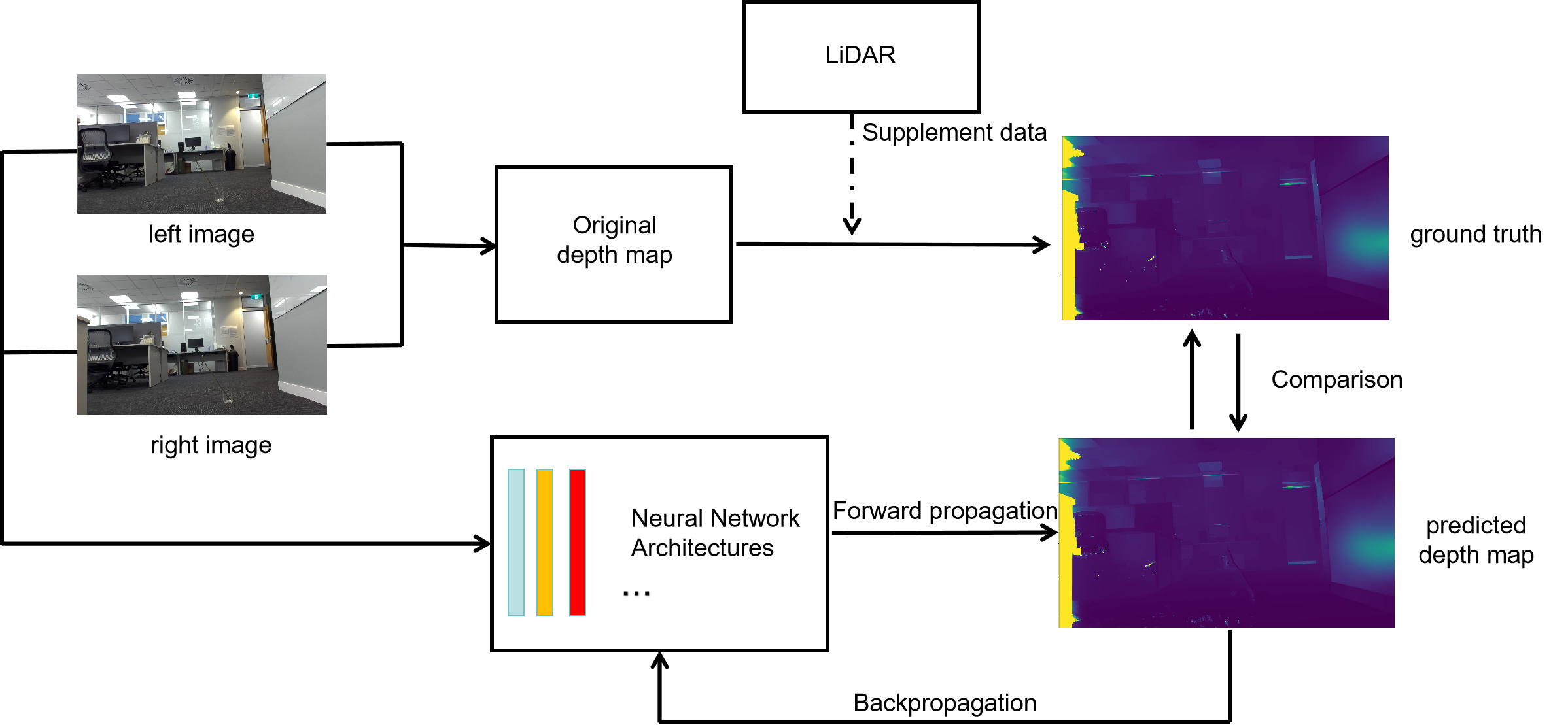}
        \caption{Overall workflow of the proposed branch-positioning pipeline. The
        rectified stereo pair is fed in parallel to a segmentation network and a
        stereo disparity estimator; the two outputs are fused through centroid-based
        triangulation and MAD outlier rejection to yield a single branch-to-camera
        distance.}
        \label{fig:pipeline_overview}
    \end{figure}

    Figure~\ref{fig:pipeline_overview} summarises the overall pipeline. A
    rectified stereo pair from the ZED~Mini camera is processed in two parallel
    branches: the left image is passed through an instance segmentation network (Mask~R-CNN
    or a YOLO segmentor) to obtain a binary branch mask, and the same stereo
    pair is passed through a disparity estimator (SGBM with WLS filtering, or
    one of the deep stereo networks) to obtain a dense disparity map. The two
    outputs are then fused: the branch mask is sampled by a centroid-based triangulation
    procedure, the sampled depth values are filtered with MAD-based outlier
    rejection, and the surviving values are averaged into a single robust distance
    estimate. Each block is described in turn in Sections~\ref{sec:method-seg}--\ref{sec:method-fusion}.

    \subsection{Branch Segmentation}
    \label{sec:method-seg}

    The segmentation stage takes the left image of the stereo pair and returns one
    binary mask per detected branch. Two model families are evaluated, motivated
    in Section~2.1: (i) Mask~R-CNN with seven backbones (ResNet-50/101 and ResNeXt-101
    in C4, DC5, and FPN configurations) as a high-mask-quality two-stage
    baseline, and (ii) the YOLOv8 (n/s/m/l/x) and YOLOv9 (c/e) segmentation variants
    as representative real-time one-stage segmentors. All models output a pixel-level
    mask which is the only quantity required by the downstream fusion step; bounding
    boxes and class scores are not used further. The specific training
    configurations are reported in Section~\ref{sec:setup} and the results in
    Section~\ref{sec:results-seg}.

    \subsection{Traditional Disparity Map Generation}
    \label{sec:method-trad}

    The traditional baseline is built from the four canonical stages of a classical
    stereo pipeline---matching-cost computation, cost aggregation, minimum-cost selection,
    and post-processing---which together yield the disparity $d$ that, with the depth
    equation Eq.~(\ref{eq:depth_concise}), gives the per-pixel depth $z$.

    For matching cost, three formulations are considered. Let $E_{l}(x,y)$ denote
    the pixel intensity at $(x,y)$ in the left image and $E_{r}(x+d,y)$ the corresponding
    pixel in the right image at disparity $d$. Absolute Intensity Differences (AD)
    and Squared Intensity Differences (SD) are simple per-pixel costs:
    \begin{equation}
        AD(x, y, d) = |E_{l}(x, y) - E_{r}(x + d, y)|
    \end{equation}
    \begin{equation}
        SD(x, y, d) = \left( E_{l}(x, y) - E_{r}(x + d, y) \right)^{2}
    \end{equation}
    Normalised Cross-Correlation (NCC) is an area-based cost that is more robust
    to brightness differences between the two cameras:
    \begin{equation}
        \begin{split}
            NCC(x, y, d) = \frac{\sum_{(i,j) \in T}\left( E_{l}(i, j) - \mu_{L}\right)
            \left( E_{r}(i + d, j) - \mu_{R}\right)}{\sqrt{\sum_{(i,j) \in T}\left(
            E_{l}(i, j) - \mu_{L}\right)^{2}\sum_{(i,j) \in T}\left( E_{r}(i + d,
            j) - \mu_{R}\right)^{2}}}
        \end{split}
    \end{equation}
    where $\mu_{L}$ and $\mu_{R}$ denote the mean intensity values of the left
    and right image patches in the window $T$:
    \begin{equation}
        \begin{split}
            \mu_{L}= \frac{1}{|T|}\sum_{(i,j) \in T}E_{l}(i, j) \\
            \mu_{R}= \frac{1}{|T|}\sum_{(i,j) \in T}E_{r}(i + d, j)
        \end{split}
    \end{equation}

    \begin{figure}[htbp]
        \centering
        \includegraphics[width=14cm]{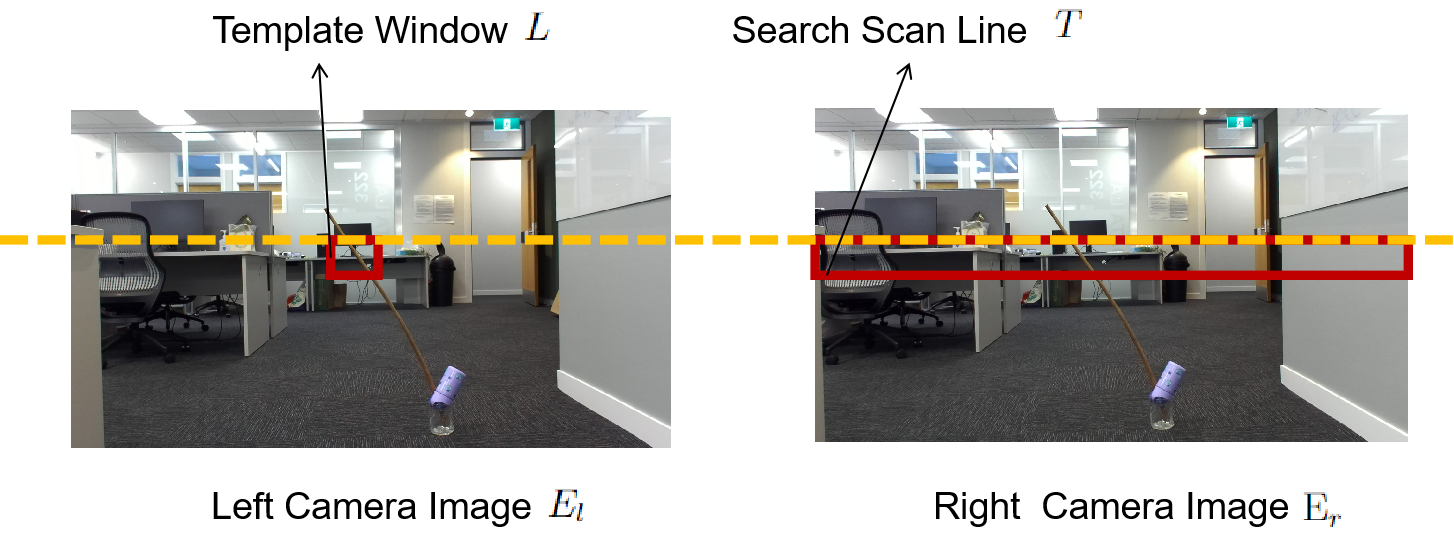}
        \caption{Block Matching illustration. $L$ denotes the template window and
        $T$ the search scan line. The left image $E_{l}$ serves as the reference,
        and the corresponding pixel is located in $E_{r}$.}
        \label{fig:block_matching}
    \end{figure}

    Cost aggregation combines costs from the local neighbourhood. The fixed
    window aggregator sums costs in a single window of size $|T|$:
    \begin{equation}
        C_{fixed}(x, y, d) = \sum_{(i,j) \in T}C(i, j, d)
    \end{equation}
    The multiple-window variant considers $N$ shifted windows $T_{k}$:
    \begin{equation}
        C_{multi}(x, y, d) = \sum_{k=1}^{N}\sum_{(i,j) \in T_k}C(i, j, d)
    \end{equation}
    Iterative diffusion propagates costs across the neighbourhood over $N$
    iterations with weights $w_{n}$:
    \begin{equation}
        C_{diff}(x, y, d) = \sum_{n=1}^{N}w_{n}\cdot C_{diff}^{(n-1)}(x, y, d)
    \end{equation}

    Disparity is then selected pixel-by-pixel as the cost minimum (local methods)
    or by minimising a global energy that combines a data term and a smoothness
    term controlled by $\lambda$:
    \begin{equation}
        d(x, y) = \arg \min_{d}C(x, y, d)
    \end{equation}
    \begin{equation}
        E(d) = E_{data}(d) + \lambda E_{smooth}(d)
    \end{equation}
    \begin{equation}
        E_{data}(d) = \sum_{(x,y)}C(x, y, d(x, y))
    \end{equation}
    \begin{equation}
        \begin{split}
            E_{smooth}(d) = \sum_{(x,y)}\rho(d(x, y) - d(x+1, y)) + \rho(d(x, y)
            - d(x, y+1))
        \end{split}
    \end{equation}

    Finally, post-processing refines the raw disparity through left-right consistency
    checking, sub-pixel interpolation,
    \begin{equation}
        \begin{split}
            d_{sub}(x, y) = d(x, y) - \frac{C(x, y, d+1) - C(x, y, d-1)}{2 \left(
            C(x, y, d+1) + C(x, y, d-1) - 2C(x, y, d) \right)}
        \end{split}
    \end{equation}
    and median-style filtering,
    \begin{equation}
        d_{post}(x, y) = \text{median}\{ d(x+i, y+j) | (i, j) \in W \}
    \end{equation}

    On top of this classical post-processing, weighted least squares (WLS)
    filtering is applied as an edge-preserving smoother. Letting $d'$ denote the
    optimised disparity, $w_{(x,y)}$ a per-pixel data weight, $w_{(x,y),(x',y')}$
    an inter-pixel similarity weight, $\sigma$ a colour smoothness parameter, and
    $\lambda$ the data--smoothness balance, WLS minimises
    \begin{equation}
        \begin{split}
            d' = \arg \min_{d}\Bigg(\sum_{(x,y)}w_{(x,y)}(d'(x,y) - d(x,y))^{2}+
            \lambda\sum_{(x,y),(x',y')}w_{(x,y),(x',y')}(d'(x,y) - d'(x',y'))^{2}
            \Bigg)
        \end{split}
    \end{equation}
    with similarity weights computed in image space:
    \begin{equation}
        w_{(x,y),(x',y')}= \exp \left( - \frac{(E_{l}(x,y) - E_{l}(x',y'))^{2}}{2\sigma^{2}}
        \right)
    \end{equation}

    The Semi-Global Block Matching (SGBM) algorithm is used as the instantiation
    of this generic pipeline in the experiments. A worked example of every stage
    on a representative branch image is shown in the results section (Figure~\ref{fig:sgbm_pipeline}).

    \subsection{Deep-Learning Disparity Map Generation}
    \label{sec:method-deep}

    Six deep stereo networks are evaluated as drop-in replacements for the SGBM block
    in Figure~\ref{fig:pipeline_overview}: PSMNet, ACVNet, GWCNet, MobileStereoNet,
    RAFT-Stereo, and NeRF-Supervised Deep Stereo. Their architectural details
    and selection rationale are summarised in Section~2.2. From the pipeline's point
    of view, all six produce a dense per-pixel disparity map of the same size as
    the input image, and therefore can be plugged into the same downstream fusion
    step described next without modification. Two monocular networks (MiDaS and Depth
    Anything) are also reported, but only as a sanity check; they cannot produce
    metric depth without an external scale and are therefore not integrated into
    the main pipeline.

    \subsection{Branch-to-Camera Distance via Centroid Triangulation and MAD
    Filtering}
    \label{sec:method-fusion}

    The fusion stage is the main methodological contribution of this paper.
    Given a binary branch mask from Section~\ref{sec:method-seg} and a disparity
    (and hence depth) map from Section~\ref{sec:method-trad} or~\ref{sec:method-deep},
    the goal is to return a single robust branch-to-camera distance, even when
    individual disparity pixels inside the mask are noisy or grossly wrong.

\begin{figure}[htbp]
        \centering
        \subfigure[predicted points]{\includegraphics[width=0.45\textwidth]{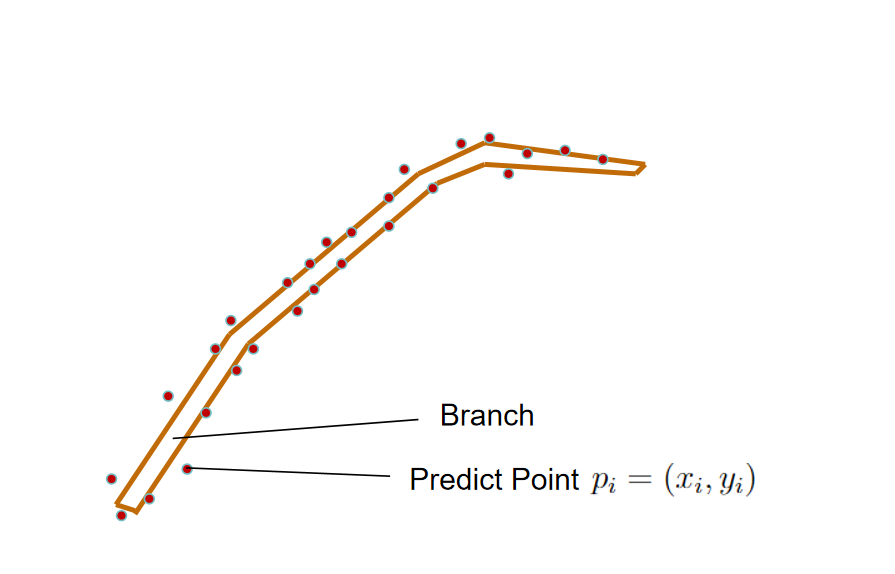}}
        \hfill \subfigure[grouping the closest points into triangles]{\includegraphics[width=0.35\textwidth]{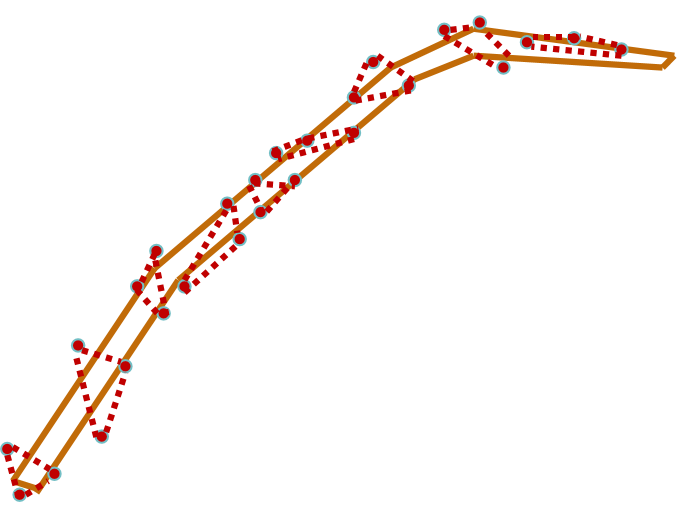}}
        \vspace{\baselineskip}
        \subfigure[computing centroids]{\includegraphics[width=0.45\textwidth]{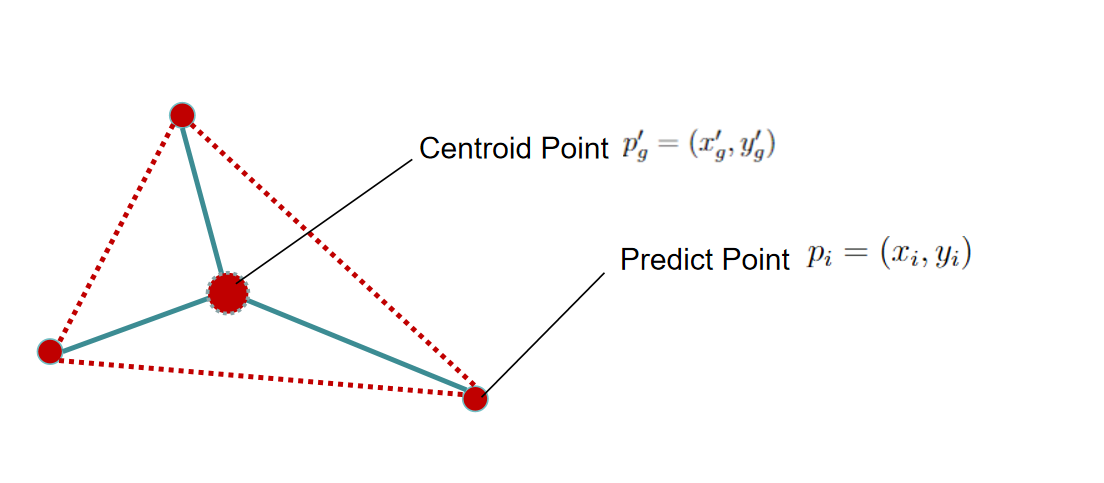}}
        \caption{Centroid-based branch localisation: (a) predicted boundary points
        around the branch, (b) the closest points grouped into triangles, and (c)
        the resulting centroid sample locations.}
        \label{fig:centroid_process}
    \end{figure}

    \paragraph{Centroid-based sampling.}
    Let $P$ denote the set of $n$ predicted boundary points around a branch,
    $P=\{ p_{1}, p_{2},\ldots, p_{n}\}$ with $p_{i}=(x_{i},y_{i})$ in image
    coordinates. Rather than sampling depth at every mask pixel, the proposed
    procedure groups the boundary points into triangles and uses each triangle's
    centroid as a sampling location. This concentrates samples on the medial line
    of the branch, where stereo matching is most reliable. Let the centroid set be
    $P'= \{p'_{1}, p'_{2},\ldots,p'_{K}\}$, $0<K\leq \lfloor n/3 \rfloor$, with $p
    '_{g}= (x'_{g}, y'_{g})$. Figure~\ref{fig:centroid_process} illustrates the
    construction.

    \paragraph{Neighbourhood expansion.}
    Each centroid is expanded by $m$ neighbouring sample points to reduce the
    variance of the depth estimate. The $j$-th expanded point around centroid $p'
    _{i}$ is denoted $q_{i,j}= (x''_{i,j}, y''_{i,j})$. Defining
    \[
        P''_{i}= \{q_{i,1}, q_{i,2}, \ldots, q_{i,m}\}, \qquad P'' = \{P''_{1}, P
        ''_{2}, \ldots, P''_{K}\},
    \]
    the full pool of candidate sample locations is $P''' = P'' \cup P'$.

    \paragraph{MAD-based outlier rejection.}
    Let $D(P''')$ denote the multiset of depth values read from the depth map at
    the locations in $P'''$. Outliers are removed using the Median Absolute
    Deviation,
    \begin{equation}
        \text{MAD}= \text{median}\bigl(\bigl|D(P''') - \text{median}(D(P'''))\bigr
        |\bigr)
    \end{equation}
    A sample is retained if its depth lies within $\text{median}(D(P''')) \pm k \cdot
    \text{MAD}$, where $k$ is a user-specified rejection threshold (set to $k=3$
    in the experiments). The retained set $\bar{P}$ yields the final distance:
    \begin{equation}
        \text{distance}= \text{mean}(\bar{P}) \label{eq:distance}
    \end{equation}
    A worked example of the resulting samples on the depth map and on the RGB
    image is shown in Figure~\ref{fig:centroid_depth} of Section~\ref{sec:results}.

    \paragraph{Why this resolves the research gap.}
    Standard mean- or median-of-mask approaches use every pixel inside the
    branch mask, which on thin branches dilutes the estimate with background-bleed
    disparities along the silhouette. Centroid sampling biases the samples
    toward the branch interior, while MAD-based rejection (rather than naive
    mean) is robust to a small but high-magnitude tail of bad disparities. Both
    choices directly target the failure modes identified in Section~\ref{sec:overview}.

    As a fallback, a polygon-based variant is also implemented: the boundary
    points are connected into an irregular polygon, all enclosed pixels are taken
    as $P'''$, and the same MAD filter and mean of Eq.~(\ref{eq:distance}) are applied.
    This variant samples many more points and is suitable for statistical
    analysis of the depth distribution along the entire branch; it is reported alongside
    the centroid variant in Section~\ref{sec:results}.

    \section{Data Collection}
    \label{sec:data}

    The dataset used in this paper consists of stereo image pairs of radiata
    pine branches captured indoors with a ZED~Mini stereo camera
    \url{https://store.stereolabs.com/products/zed-mini} at a resolution of
    $1920 \times 1080$ per view. Branches are placed at controlled distances from
    the camera (1\,m, 1.5\,m, and 2\,m) and photographed under a range of laboratory
    lighting conditions in different corners of the lab, in order to discourage
    the segmentor from over-fitting to a single illumination or background. In total,
    71 stereo pairs are captured: 61 pairs (122 images) form the training set and
    the remaining 10 pairs the test set.

    The annotation protocol marks key boundary points around each branch so that
    the resulting polygon completely encloses the branch silhouette; these
    polygons serve both as segmentation ground truth and as the boundary point
    set $P$ used by the centroid procedure of Section~\ref{sec:method-fusion}.

    Indoor data collection is a deliberate compromise. It allows accurate control
    of branch--camera distance (and therefore an approximate distance reference for
    the depth experiments), repeatable lighting, and fast iteration on the
    perception pipeline, but it does not capture the wind motion, illumination
    changes, and dense canopy clutter of an operational forestry setting. The
    dataset is also relatively small; however, because the segmentation task in
    this paper involves a single object class with a comparatively simple
    silhouette, modern segmentors nevertheless reach near-saturated accuracy on it
    (see Section~\ref{sec:results-seg}). Field collection from the drone
    platform itself, with corresponding LiDAR-based reference depths, is
    identified as the most important dataset extension and is discussed further
    in Section~\ref{sec:future}.

    \section{Experimental Setup}
    \label{sec:setup}

    \subsection{Datasets}
    All segmentation models are trained on the COCO dataset \citep{lin2014microsoft}
    (for the COCO-baseline columns) and on the 71-pair branch dataset described
    in Section~\ref{sec:data} (for the Branches columns). For the deep stereo networks,
    three public stereo datasets are used: KITTI~2012, KITTI~2015, and Scene~Flow.
    KITTI~2012 and KITTI~2015 contain rectified stereo pairs of urban driving
    scenes with high-quality LiDAR-derived ground-truth disparity; Scene~Flow contains
    synthetically rendered stereo pairs with dense ground-truth disparity. These
    three datasets are used both as pre-training sources and to study cross-dataset
    transfer to the branch domain.

    \subsection{Training Configuration}
    All segmentation models are trained for 100 epochs each, a budget chosen to keep
    the comparison balanced given the small size of the branch dataset. PSMNet
    is fine-tuned for 100 epochs on KITTI~2012 and 100 epochs on KITTI~2015 in
    the cross-dataset transfer study. Default optimiser, learning-rate schedule,
    and data-augmentation settings are used for each model from the corresponding
    reference implementation; no model-specific hyperparameter search is
    performed. The MAD rejection threshold in Eq.~(\ref{eq:distance}) is set to
    $k=3$, and the neighbourhood expansion size to $m=8$, in all experiments.

    \subsection{Evaluation Metrics}
    \label{sec:metrics}

    \paragraph{Segmentation.}
    Detection and segmentation quality are measured with the standard COCO mean
    Average Precision metric, averaged over IoU thresholds from $0.5$ to $0.95$ in
    steps of $0.05$ \citep{lin2014microsoft}:
    \begin{equation}
        \text{mAP}_{50\text{--}95}= \frac{1}{n}\sum_{t=0.5}^{0.95}\text{AP}(t), \qquad
        n=10 \label{eq:mAP}
    \end{equation}
    where $AP(t)$ denotes the standard Average Precision at IoU threshold $t$.
    Both bounding-box and mask versions (mAP\textsubscript{box50--95}, mAP\textsubscript{mask50--95})
    are reported.

    \paragraph{Disparity / depth.}
    On datasets where reference disparity is available, disparity quality is reported
    using the Root Mean Squared Error,
    \begin{equation}
        \text{RMSE}= \sqrt{\frac{1}{N}\sum_{i=1}^{N}(y_{i}- \hat{y}_{i})^{2}}, \label{eq:rmse}
    \end{equation}
    where $y_{i}$ and $\hat{y}_{i}$ are the reference and predicted values and
    $N$ is the number of evaluated pixels. On the branch dataset, where dense
    reference disparity is unavailable, depth quality is assessed qualitatively (Figures~\ref{fig:sgbm_pipeline}
    and~\ref{fig:depth_comparison}) and through histograms of the predicted distance
    distribution at known camera-to-branch distances of 1\,m, 1.5\,m, and 2\,m (Figure~\ref{fig:histograms}).

    \section{Results and Discussion}
    \label{sec:results}

    The experiments test three hypotheses motivated by the introduction: (H1) modern
    instance segmentors are accurate enough on radiata pine branches that the segmentation
    stage is \emph{not} the bottleneck of the pipeline; (H2) deep stereo networks
    produce more coherent disparity on thin branches than the SGBM+WLS baseline,
    but suffer from a domain gap when trained only on urban driving data; and (H3)
    the proposed centroid-based triangulation with MAD filtering turns a noisy disparity
    map into a stable branch-to-camera distance at the working range of 1--2\,m.

    \subsection{Branch Segmentation}
    \label{sec:results-seg}

    Table~\ref{tab:yolo} reports mAP\textsubscript{box50--95} and mAP\textsubscript{mask50--95}
    for seven Mask~R-CNN backbones and seven YOLO segmentation variants on both COCO
    and the branch dataset.

    \begin{table}[h!]
        \centering
        \definecolor{rowgray}{gray}{0.92}
        \rowcolors{3}{white}{rowgray}
        \begin{tabular}{lccccc}
            \toprule \textbf{Model}              & \multicolumn{2}{c}{\textbf{COCO}} & \multicolumn{2}{c}{\textbf{Branches}} \\
            \cmidrule(lr){2-3}\cmidrule(lr){4-5} & mAP\textsubscript{box50--95}      & mAP\textsubscript{mask50--95}        & mAP\textsubscript{box50--95} & mAP\textsubscript{mask50--95} \\
            \midrule Mask R-CNN R50-C4           & 39.8                              & 34.4                                 & 76.86                        & 0.06                          \\
            Mask R-CNN R50-DC5                   & 40.0                              & 35.9                                 & 77.54                        & 9.16                          \\
            Mask R-CNN R50-FPN                   & 41.0                              & 37.2                                 & 79.19                        & 6.75                          \\
            Mask R-CNN R101-C4                   & 42.6                              & 36.7                                 & 88.05                        & 0.05                          \\
            Mask R-CNN R101-DC5                  & 41.9                              & 37.3                                 & 79.12                        & 9.94                          \\
            Mask R-CNN R101-FPN                  & 42.9                              & 38.6                                 & 84.09                        & 2.95                          \\
            Mask R-CNN X101-FPN                  & 44.3                              & 39.5                                 & 85.52                        & 11.55                         \\
            YOLOv8n-seg                          & 36.7                              & 30.5                                 & 98.9                         & 77.4                          \\
            YOLOv8s-seg                          & 44.6                              & 36.8                                 & 99.5                         & 82.0                          \\
            YOLOv8m-seg                          & 49.9                              & 40.8                                 & 99.6                         & 81.6                          \\
            YOLOv8l-seg                          & 52.3                              & 42.6                                 & 99.2                         & 80.1                          \\
            YOLOv8x-seg                          & 53.4                              & 43.4                                 & 98.7                         & 77.1                          \\
            YOLOv9c-seg                          & 52.4                              & 42.2                                 & 98.9                         & 80.9                          \\
            YOLOv9e-seg                          & 55.1                              & 44.3                                 & 98.8                         & 80.0                          \\
            \bottomrule
        \end{tabular}
        \caption{Performance comparison of Mask~R-CNN backbones and YOLOv8/v9
        segmentation variants on COCO and on the branch dataset. The C4 backbone
        follows the original Faster~R-CNN design (ResNet conv4 with a conv5 head);
        DC5 inserts dilations in conv5 (Deformable ConvNets); FPN uses ResNet with
        conv and FC heads and offers the best speed--accuracy trade-off among
        the Mask~R-CNN variants. The mAP\textsubscript{box} and mAP\textsubscript{mask}
        columns report Eq.~(\ref{eq:mAP}) for bounding-box and mask predictions,
        respectively.}
        \label{tab:yolo}
    \end{table}

    On COCO the ranking is unsurprising: heavier YOLO variants (YOLOv8x, YOLOv9e)
    lead, while smaller backbones (YOLOv8n, Mask~R-CNN~R50-C4) trail. On the branch
    dataset (evaluated on the held-out 10-pair test set described in Section~\ref{sec:data}),
    however, the picture inverts in an informative way. Mask~R-CNN's mAP\textsubscript{box50--95}
    reaches 77--88, yet its mAP\textsubscript{mask50--95} collapses to below 12,
    indicating that boxes are well placed but per-pixel masks are not tightly aligned
    with the thin branch silhouette. The YOLOv8 and YOLOv9 segmentation variants,
    in contrast, achieve mAP\textsubscript{box50--95} of 98.7--99.6 \emph{and}
    mAP\textsubscript{mask50--95} of 77.1--82.0, an absolute mask improvement of
    roughly 65--80 points over Mask~R-CNN.

    Two factors plausibly explain this gap. First, Mask~R-CNN's mask head produces
    a fixed low-resolution mask that is then upsampled to the box size, a
    strategy that is known to lose accuracy on long, thin objects; YOLO
    segmentors instead predict mask coefficients over a high-resolution
    prototype, preserving fine boundaries. Second, the branch dataset is small,
    and the heavy two-stage Mask~R-CNN heads under-fit the mask branch within
    the 100-epoch budget, while the lighter one-stage YOLO heads converge faster.

    These results support hypothesis~H1: segmentation is not the bottleneck on
    this dataset, with the YOLO variants approaching saturation on the bounding-box
    metric and reaching 77--82 mAP\textsubscript{mask50--95}. The remaining experiments
    therefore use YOLOv9e as the segmentor of choice and focus on the depth-estimation
    and fusion stages. Strengths and limitations of the dataset itself, in particular
    its limited size and indoor-only nature, mean that this near-saturation
    should not be interpreted as a solved problem in real forestry conditions; this
    is revisited in Section~\ref{sec:future}.

    \subsection{Traditional Disparity (SGBM with WLS)}
    \label{sec:results-sgbm}

\begin{figure}[htbp]
        \centering
        \subfigure[original left image]{\includegraphics[width=0.3\textwidth]{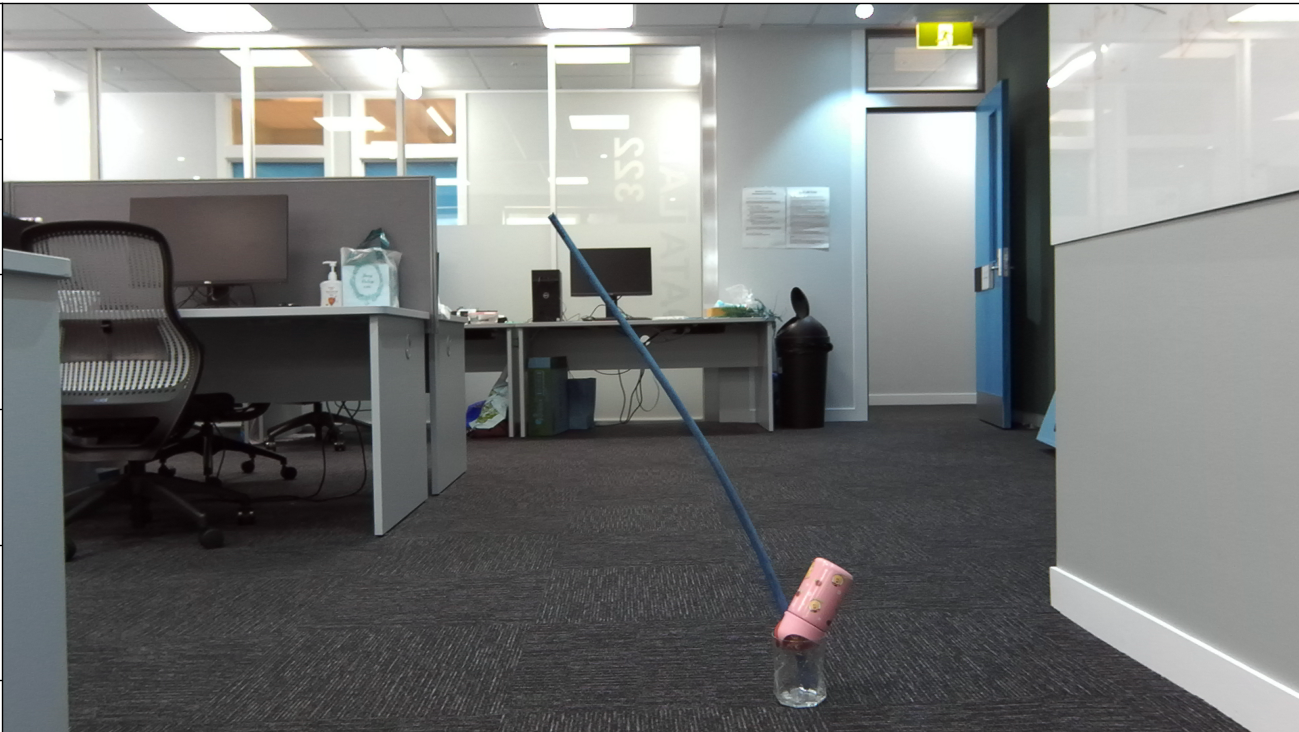}}
        \subfigure[original right image]{\includegraphics[width=0.3\textwidth]{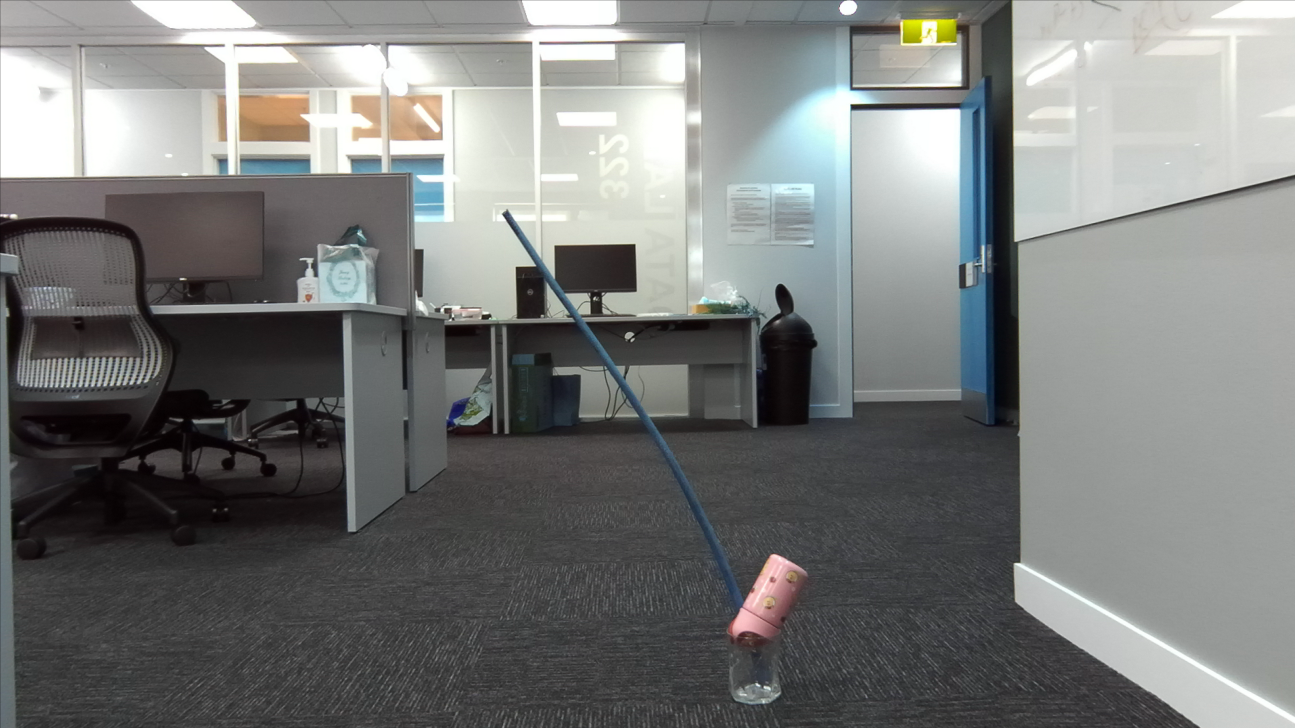}}
        \subfigure[left image after pre-processing]{\includegraphics[width=0.3\textwidth]{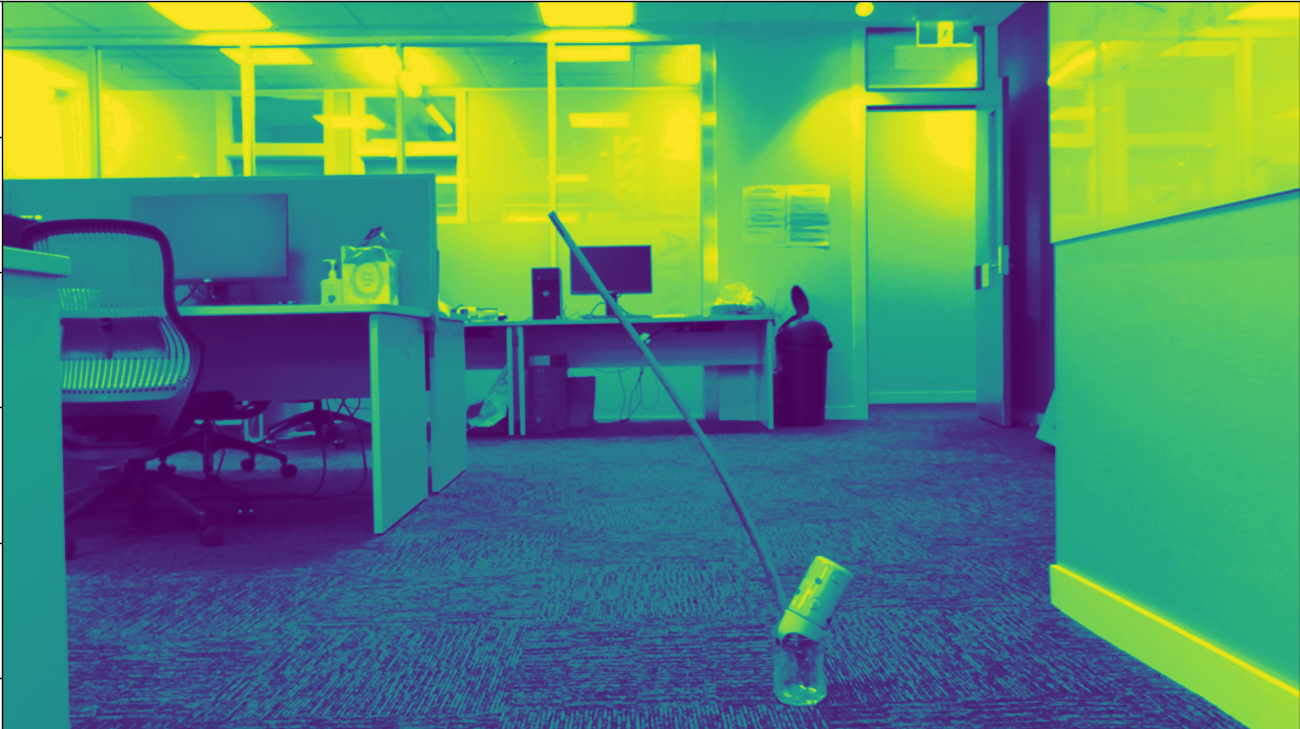}}
        \subfigure[right image after pre-processing]{\includegraphics[width=0.3\textwidth]{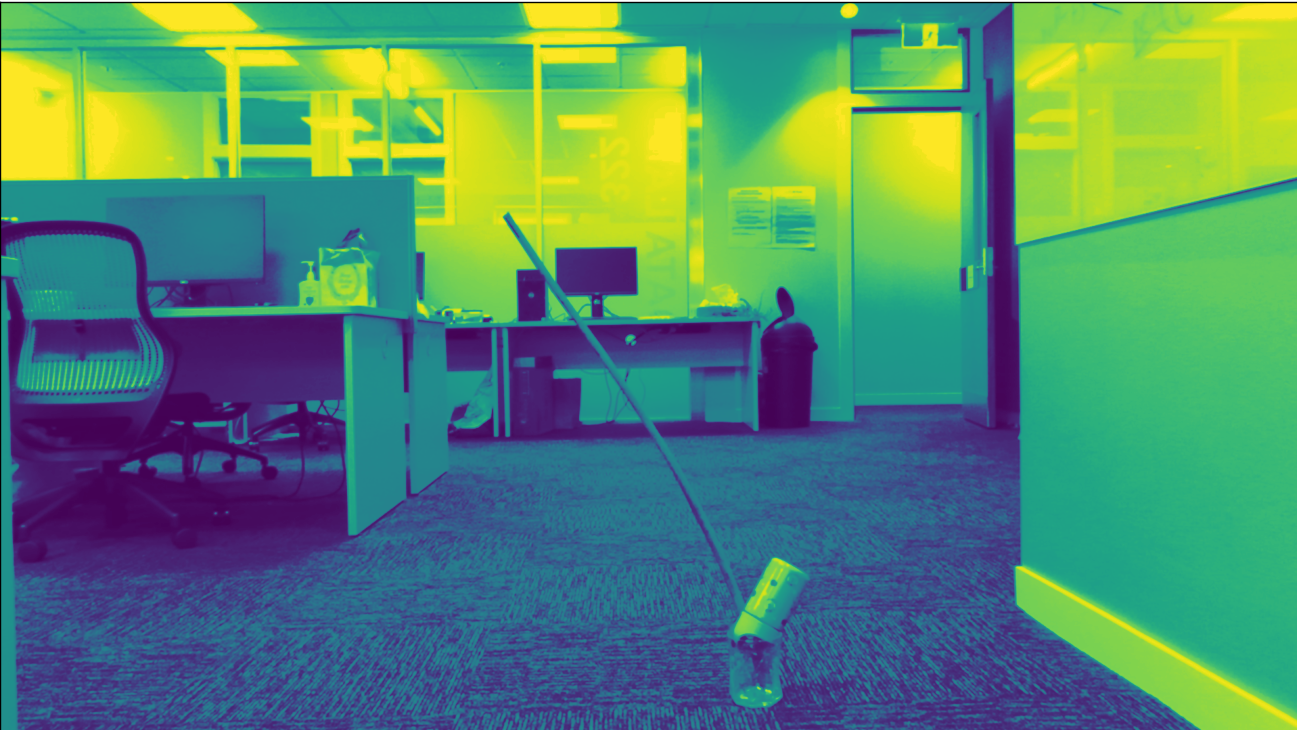}}
        \subfigure[disparity map through SGBM]{\includegraphics[width=0.3\textwidth]{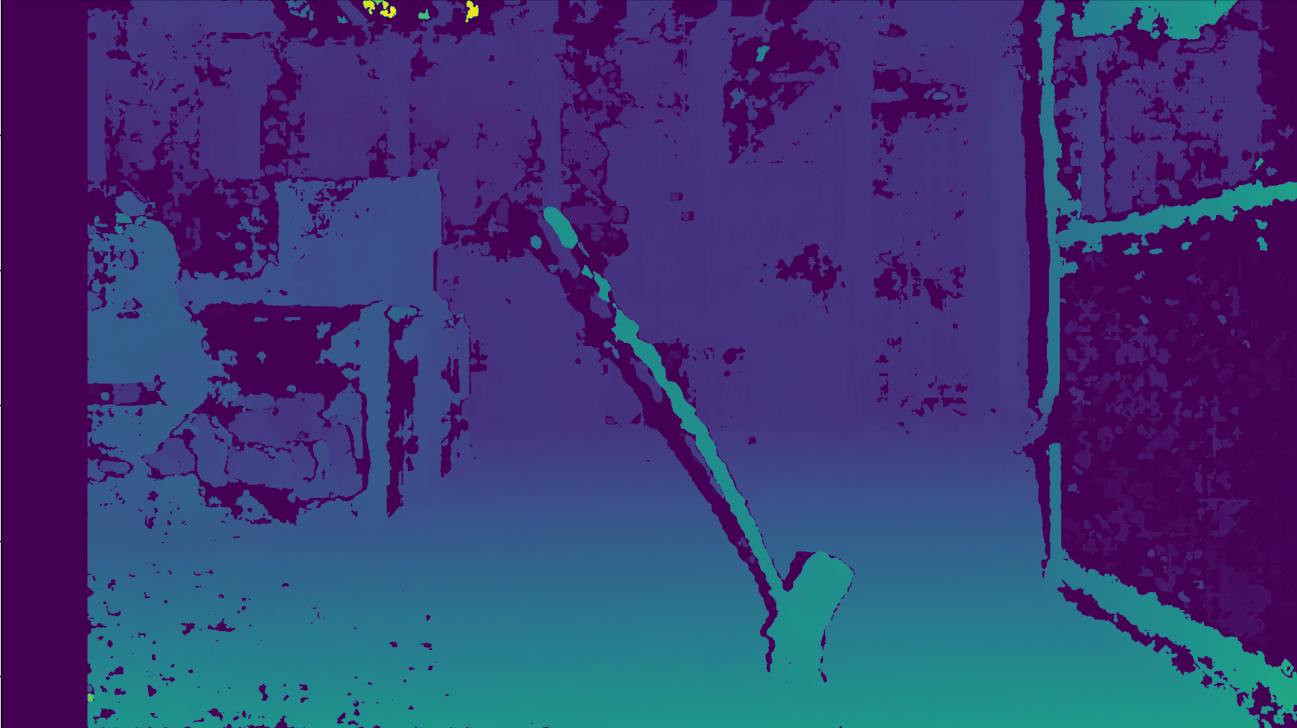}}
        \subfigure[disparity map processed by Weighted Least Squares]{\includegraphics[width=0.3\textwidth]{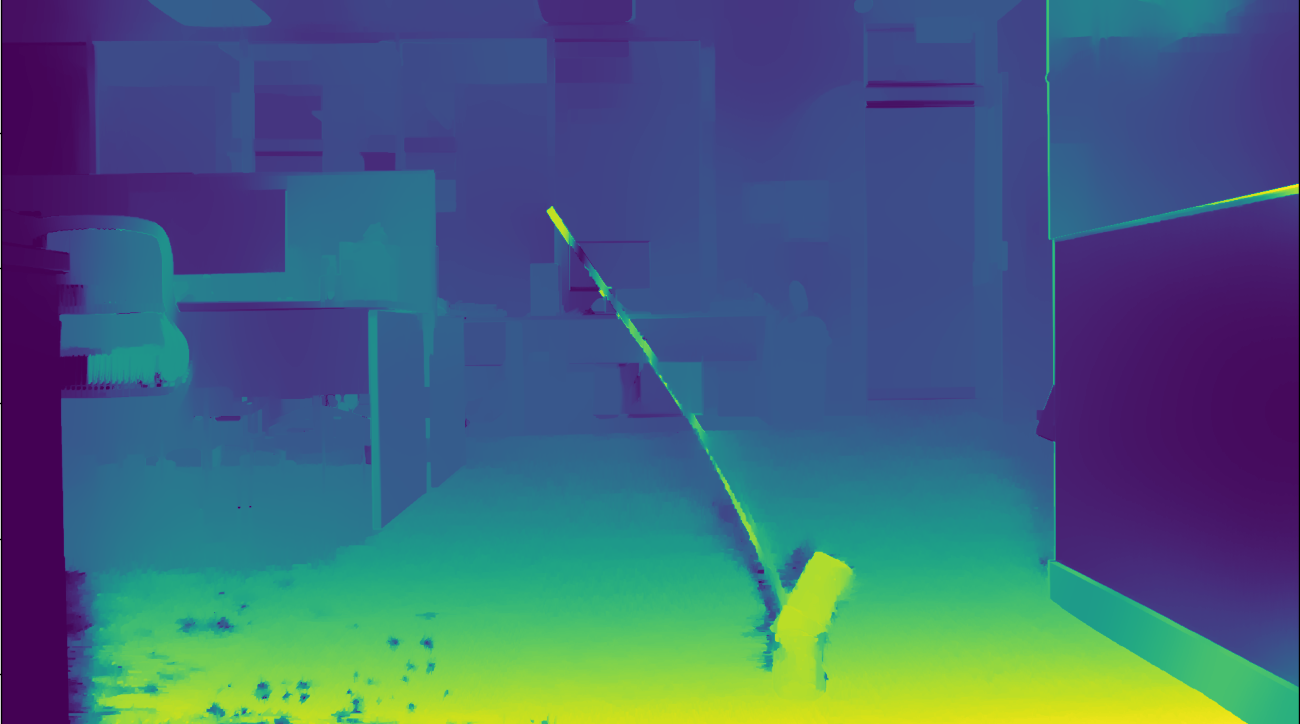}}
        \caption{SGBM disparity-map generation pipeline: (a,\,b) original left
        and right images, (c,\,d) pre-processed images, (e) SGBM-generated disparity
        map, and (f) WLS-refined disparity map.}
        \label{fig:sgbm_pipeline}
    \end{figure}

    Figure~\ref{fig:sgbm_pipeline} shows the SGBM with WLS pipeline applied to a
    representative stereo pair from the branch dataset. The raw SGBM output (e) is
    sharp on textured surfaces but exhibits the well-known failure modes of local
    stereo on thin branches: streaking artefacts, background bleed, and missing
    pixels along the silhouette. Adding WLS post-filtering (f) propagates valid disparities
    along strong image edges, visibly cleaning up the branch contour while
    preserving overall structure. SGBM with WLS therefore serves as a non-trivial
    baseline---rather than a strawman---for the deep methods that follow.

    \subsection{Monocular vs Stereo Depth (Sanity Check)}

\begin{figure}[htbp]
        \centering
        \subfigure[MiDaS, 1\,m]{\includegraphics[width=0.3\textwidth]{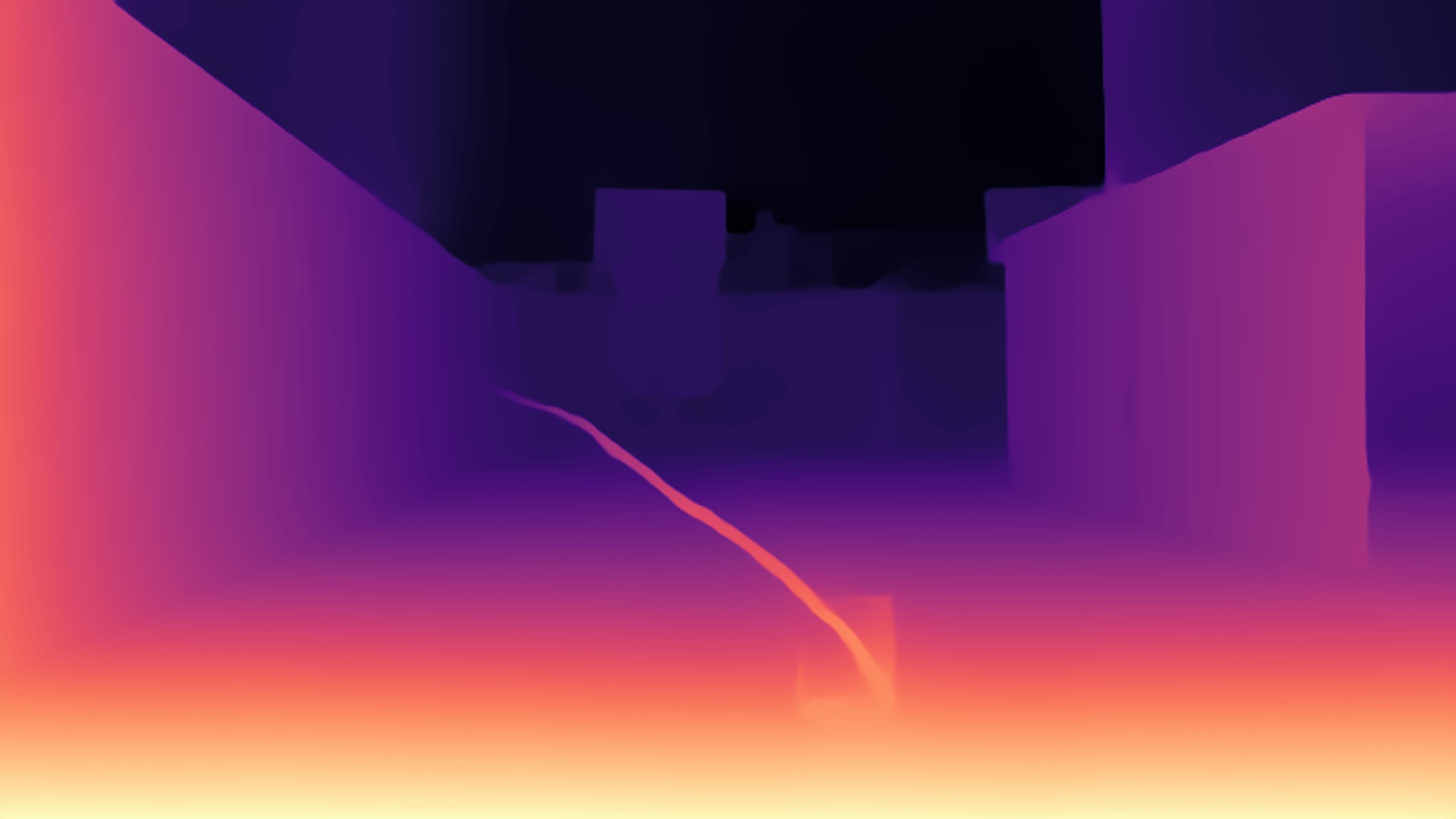}}
        \subfigure[Depth Anything, 1\,m]{\includegraphics[width=0.3\textwidth]{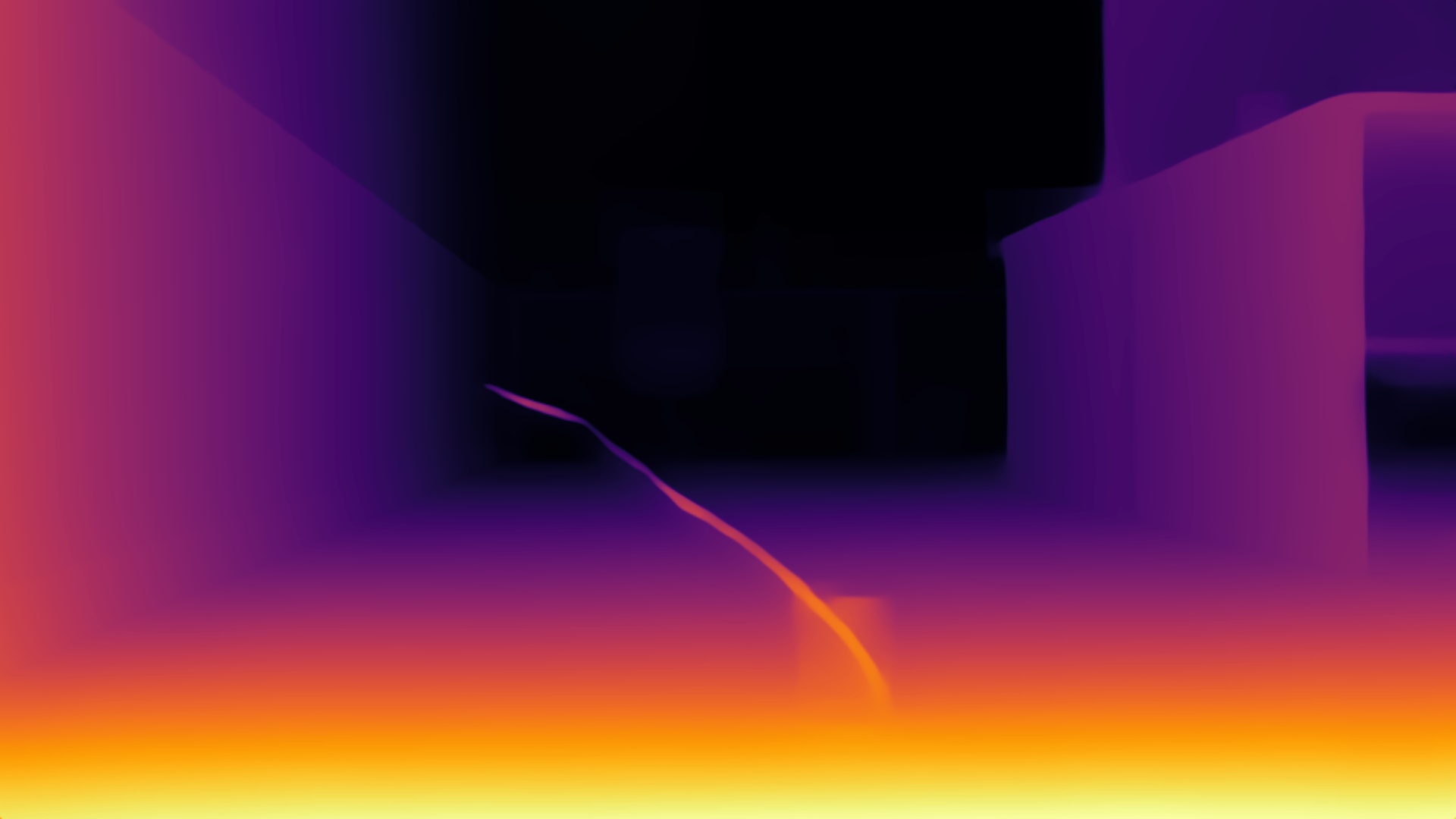}}
        \subfigure[MiDaS, 1.5\,m]{\includegraphics[width=0.3\textwidth]{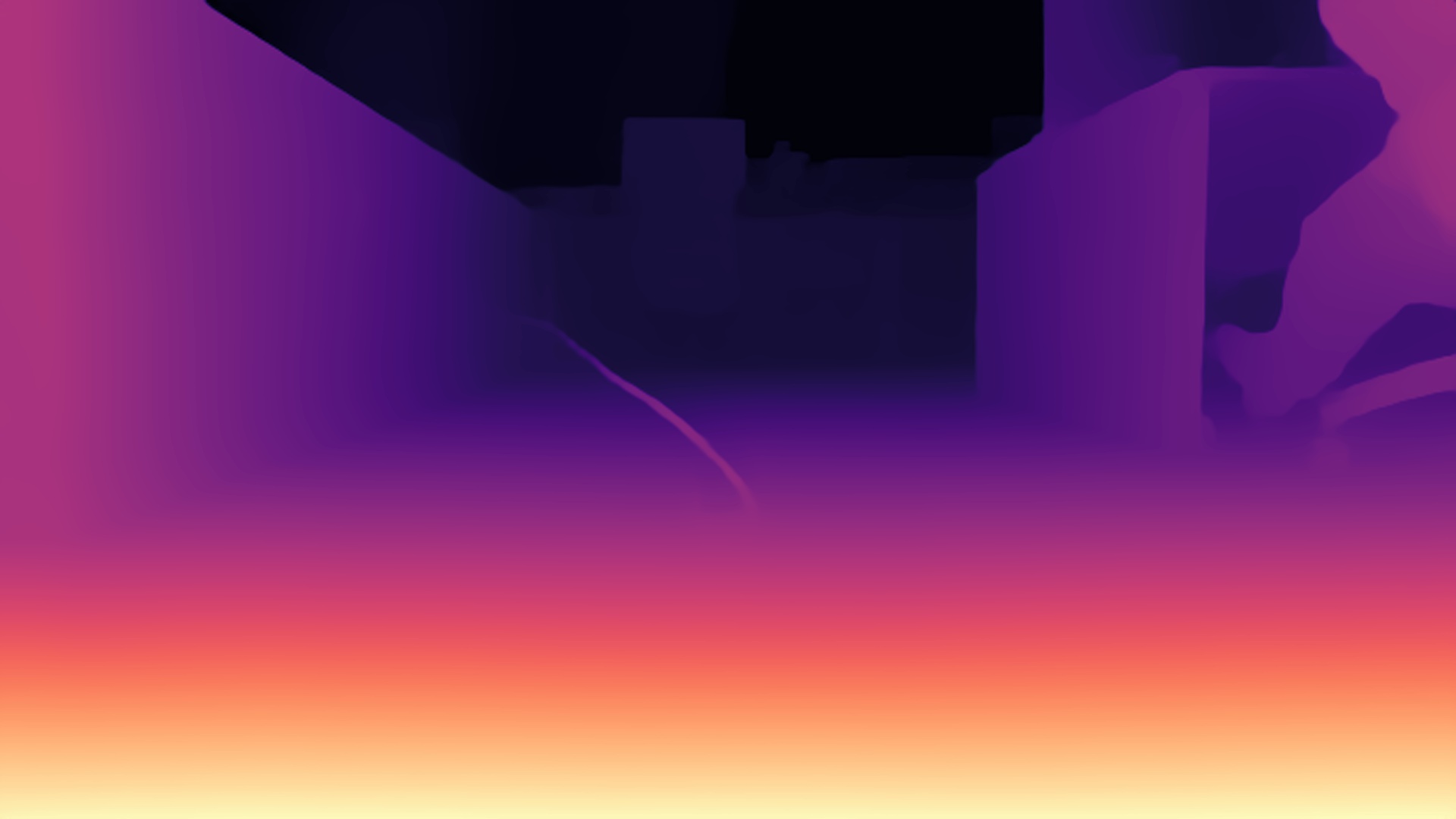}}
        \subfigure[Depth Anything, 1.5\,m]{\includegraphics[width=0.3\textwidth]{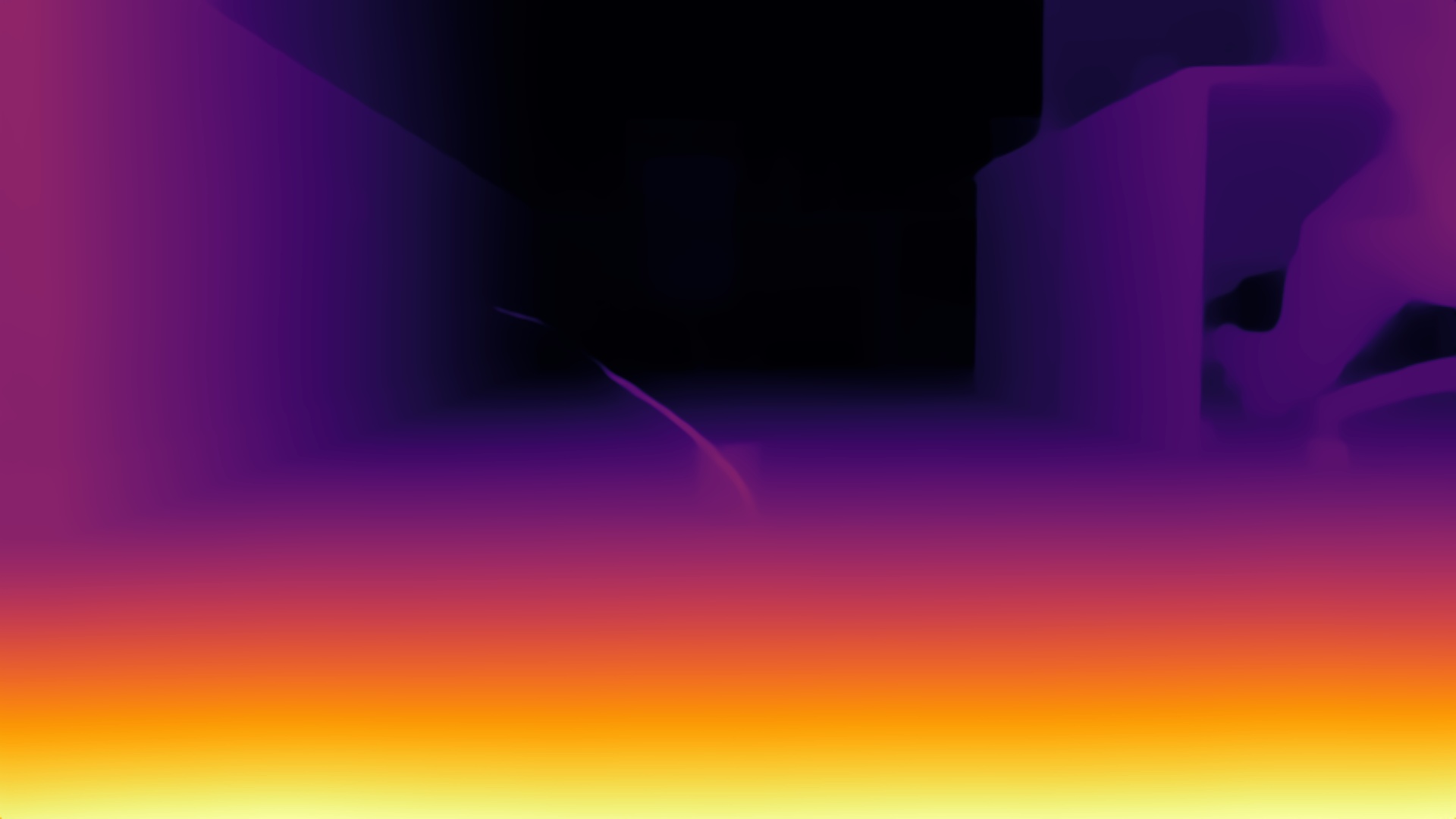}}
        \vspace{\baselineskip}
        \subfigure[MiDaS, 2\,m]{\includegraphics[width=0.3\textwidth]{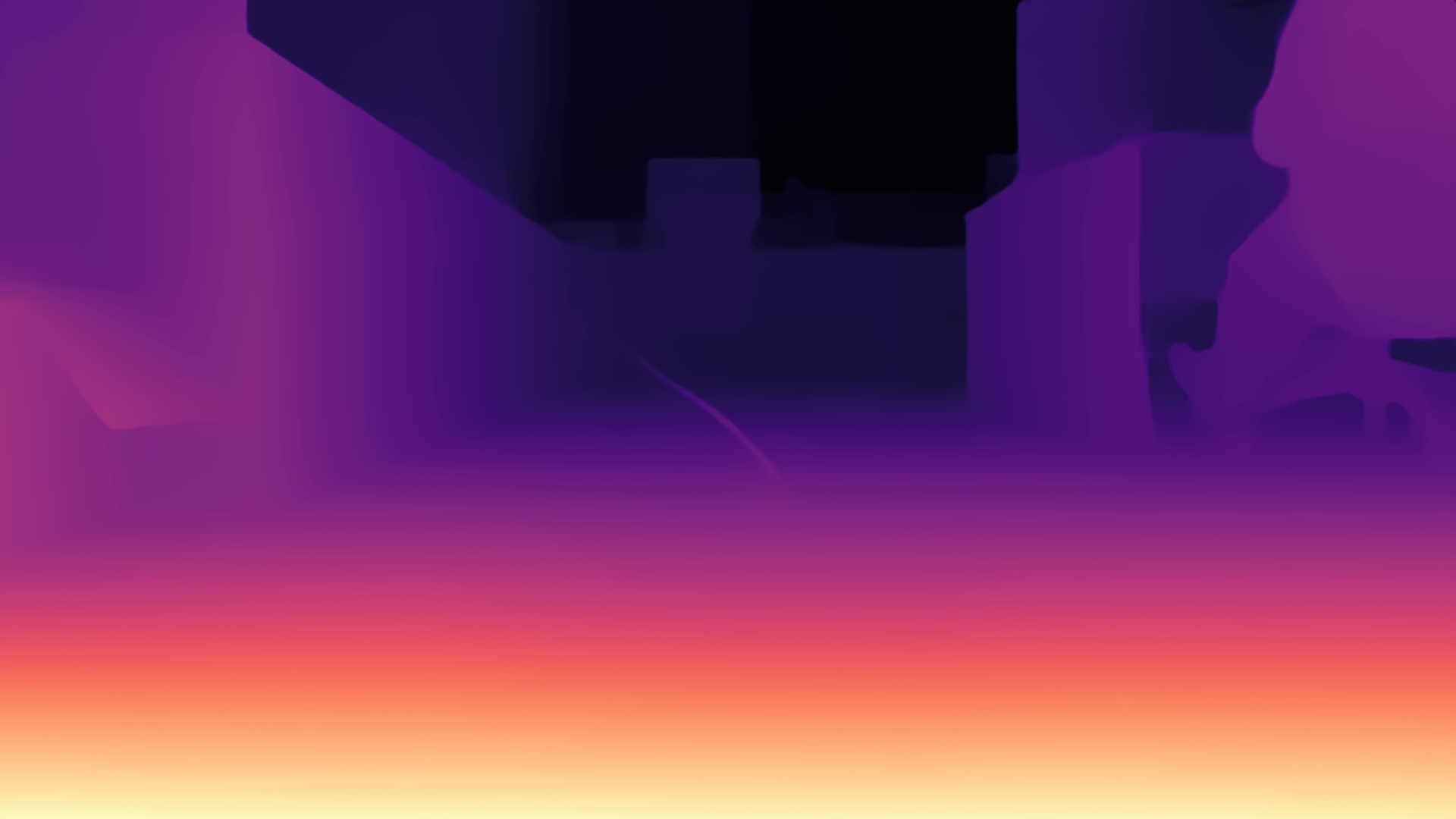}}
        \subfigure[Depth Anything, 2\,m]{\includegraphics[width=0.3\textwidth]{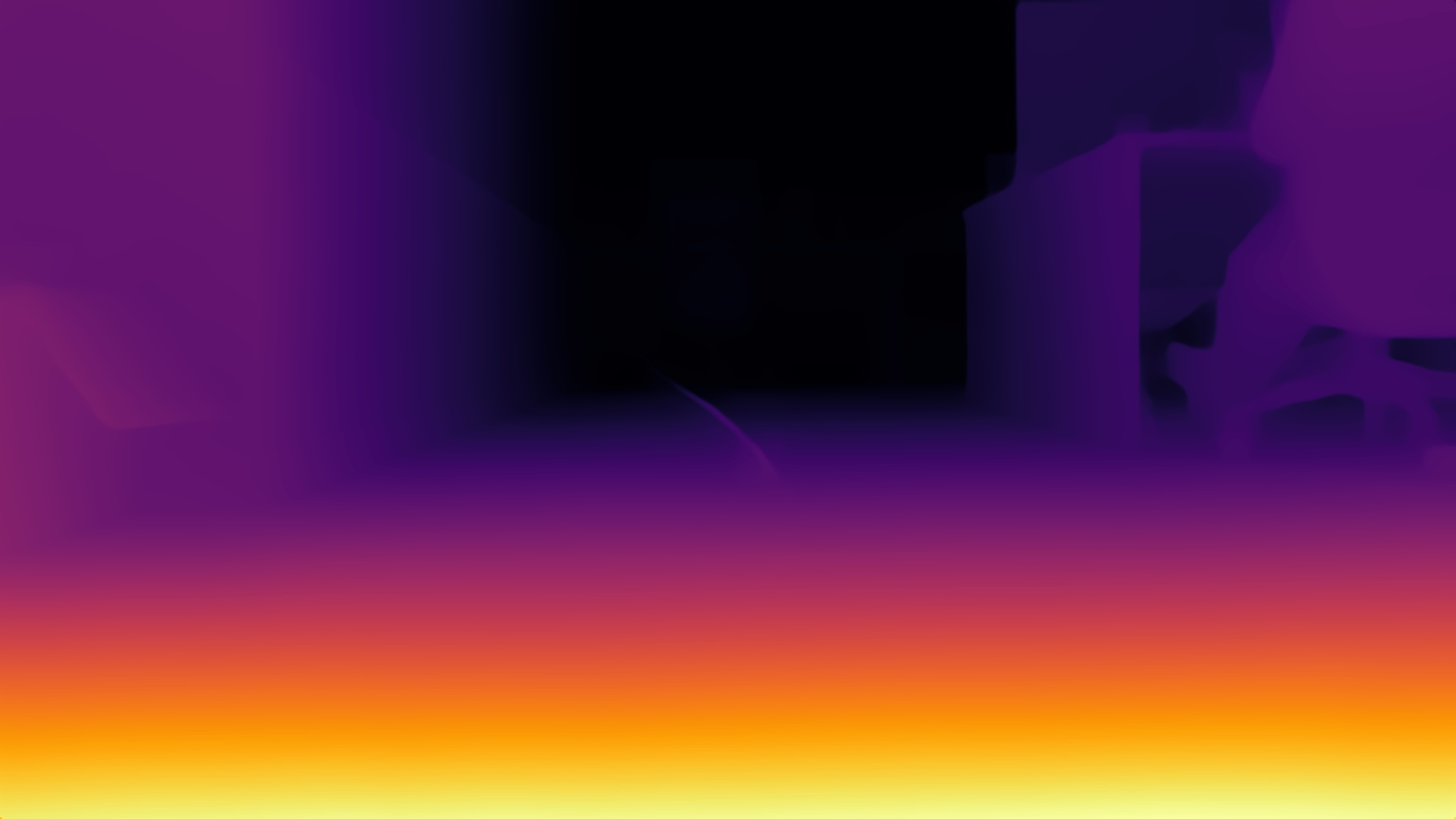}}
        \caption{Depth maps generated by MiDaS and Depth Anything at branch--camera
        distances of 1\,m, 1.5\,m, and 2\,m. Note the lack of metric scale
        change between rows.}
        \label{fig:monocular_comparison}
    \end{figure}

    Figure~\ref{fig:monocular_comparison} compares two state-of-the-art monocular
    models, MiDaS \citep{birkl2023midas} and Depth Anything \citep{yang2024depth},
    on branches at 1\,m, 1.5\,m, and 2\,m. Both networks recover the qualitative
    shape of the branch and produce plausible depth ordering, but neither yields
    metric depth: the apparent depth of the branch changes only slightly between
    1\,m and 2\,m, consistent with the well-known scale ambiguity of monocular networks.
    Depth Anything is visibly sharper on the branch silhouette than MiDaS, in line
    with prior reports, but for the present application metric depth is required,
    which motivates the use of stereo for the remainder of the experiments.

    \subsection{Cross-Dataset Fine-Tuning of Deep Stereo Networks}

    To probe the second part of hypothesis~H2, PSMNet \citep{chang2018pyramid} is
    used as a representative deep stereo network and fine-tuned for 100 epochs on
    KITTI~2012, on KITTI~2015, and pre-trained on Scene~Flow. Each variant is
    then applied to the branch dataset.

    \begin{figure}[htbp]
        \centering
        \subfigure[original left image]{\includegraphics[width=0.3\textwidth]{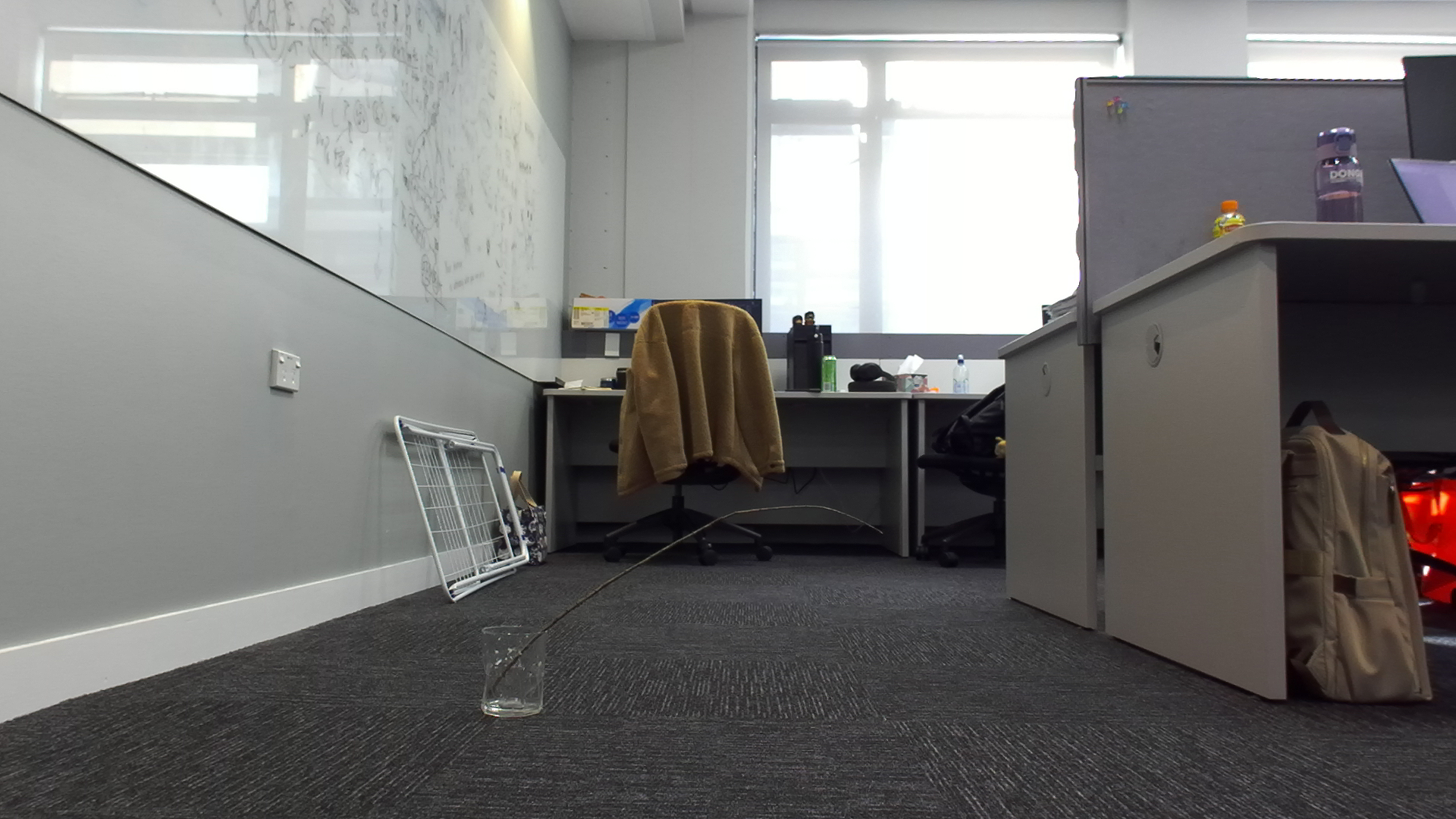}}
        \subfigure[Scene Flow pre-trained]{\includegraphics[width=0.3\textwidth]{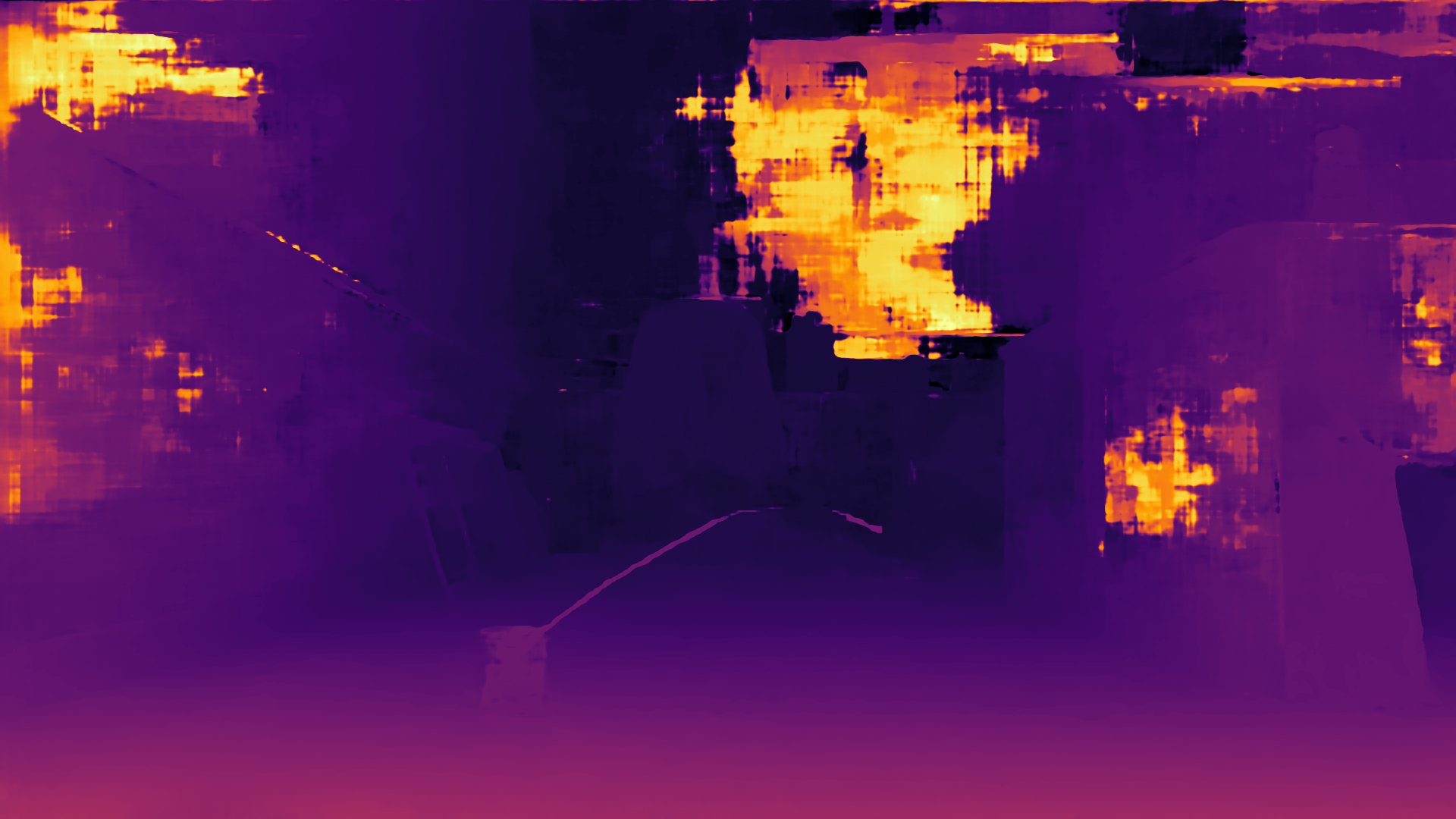}}
        \subfigure[KITTI 2012 pre-trained]{\includegraphics[width=0.3\textwidth]{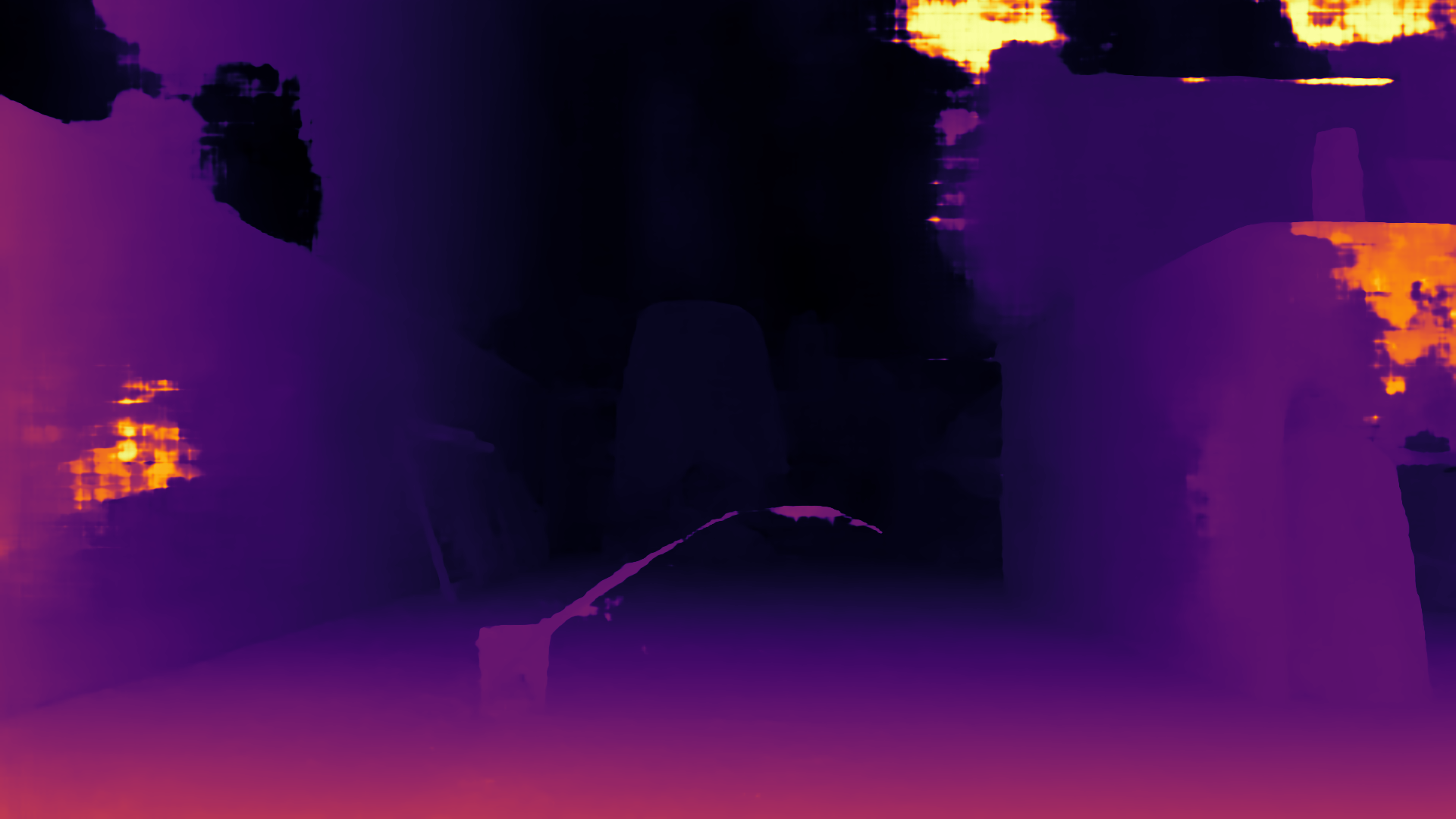}}
        \subfigure[KITTI 2012, 100 epochs]{\includegraphics[width=0.3\textwidth]{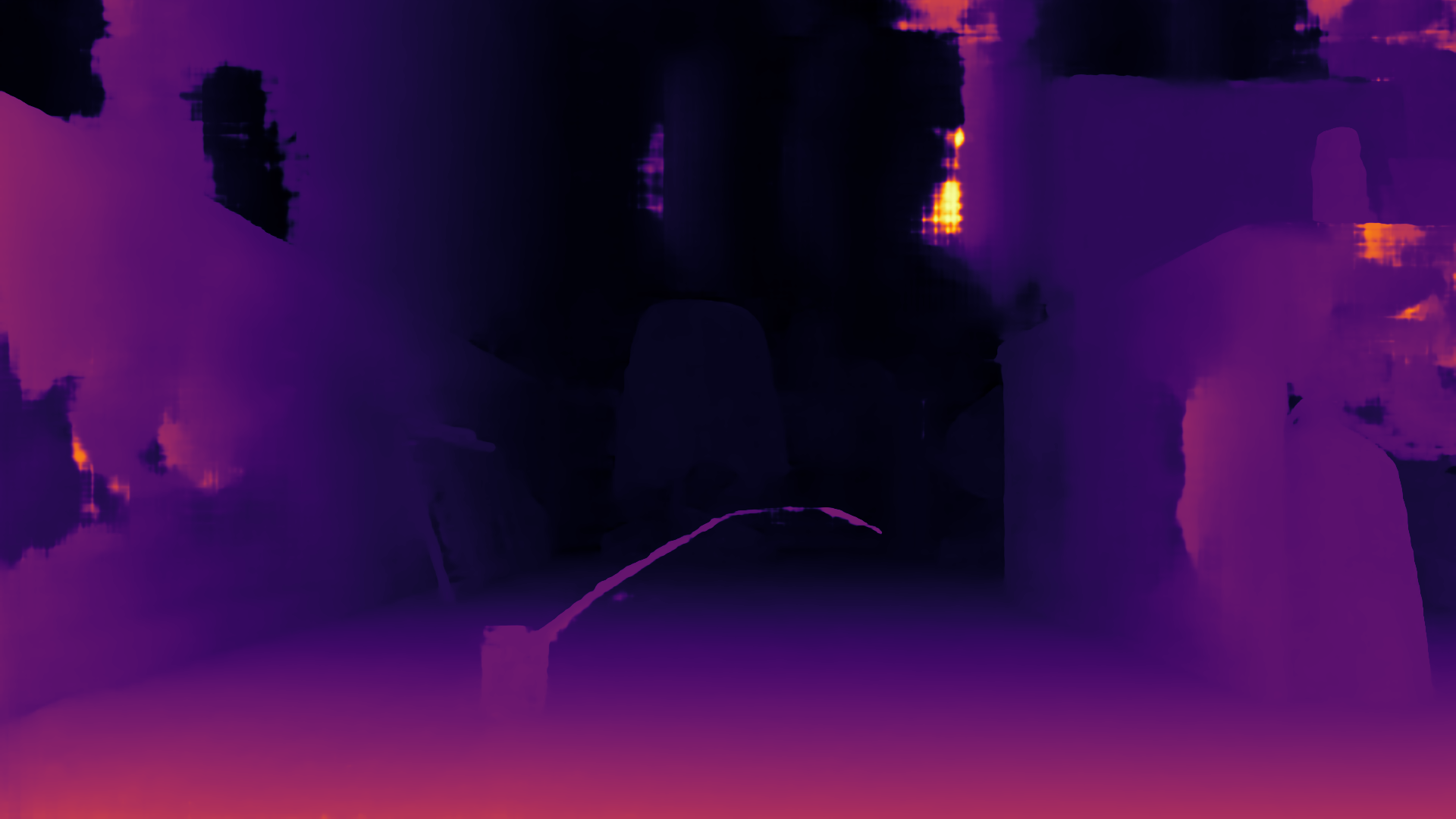}}
        \subfigure[KITTI 2015 pre-trained]{\includegraphics[width=0.3\textwidth]{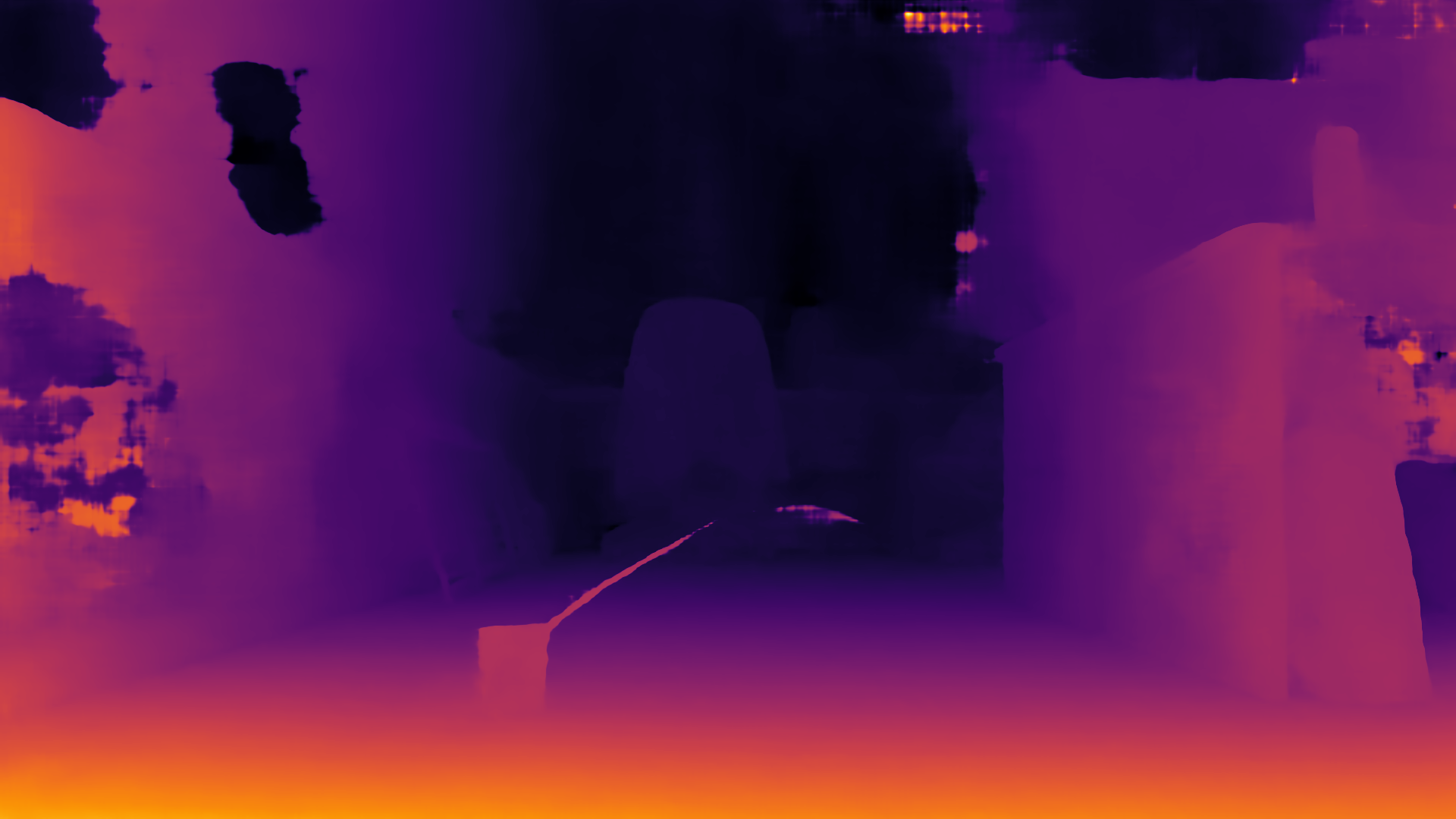}}
        \subfigure[KITTI 2015, 100 epochs]{\includegraphics[width=0.3\textwidth]{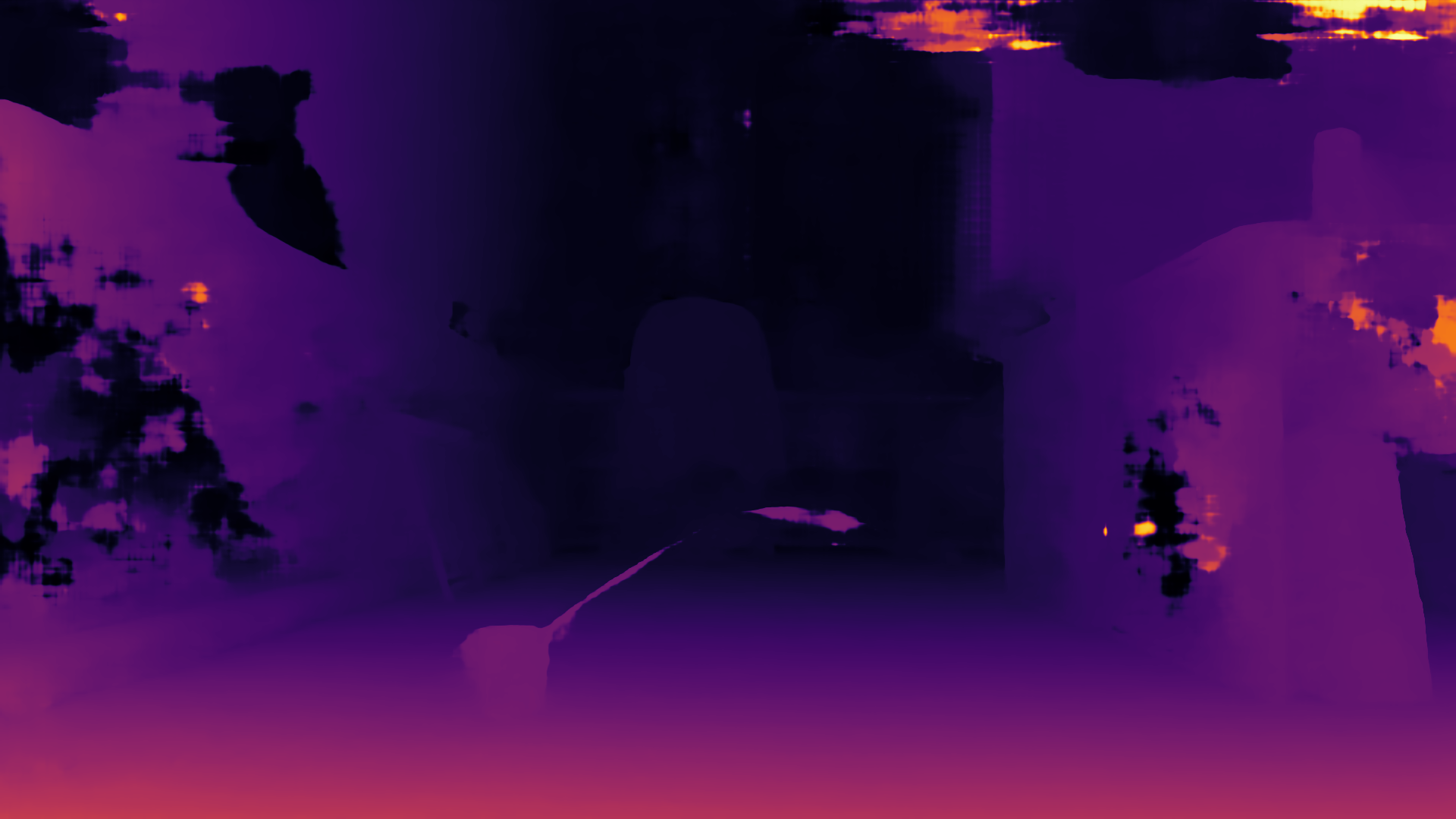}}
        \caption{PSMNet disparity on the same branch input under different pre-training
        and fine-tuning regimes.}
        \label{fig:psmnet_finetuning}
    \end{figure}

    Two patterns emerge in Figure~\ref{fig:psmnet_finetuning}. First, applying KITTI-fine-tuned
    models to branch images degrades, rather than improves, the branch disparity:
    edges become softer and the background drifts. KITTI consists of urban driving
    scenes with large, relatively textured, mostly horizontal surfaces, while
    the branch dataset contains thin, near-vertical structures with sparse natural
    texture; the KITTI-tuned model therefore appears to overfit to a different
    disparity regime. Second, the Scene~Flow pre-trained variant, which has not been
    specialised to KITTI's road geometry, retains a more usable branch disparity
    than either KITTI-fine-tuned variant. This corroborates hypothesis~H2 and motivates
    the use of pre-trained or self-/NeRF-supervised models (Scene~Flow pre-training,
    NeRF-Supervised Deep Stereo) when no in-domain ground-truth disparity is available.

    \subsection{Comparison of Deep Stereo Architectures}

\begin{figure}[htbp]
        \centering
        \subfigure[PSMNet]{\includegraphics[width=0.3\textwidth]{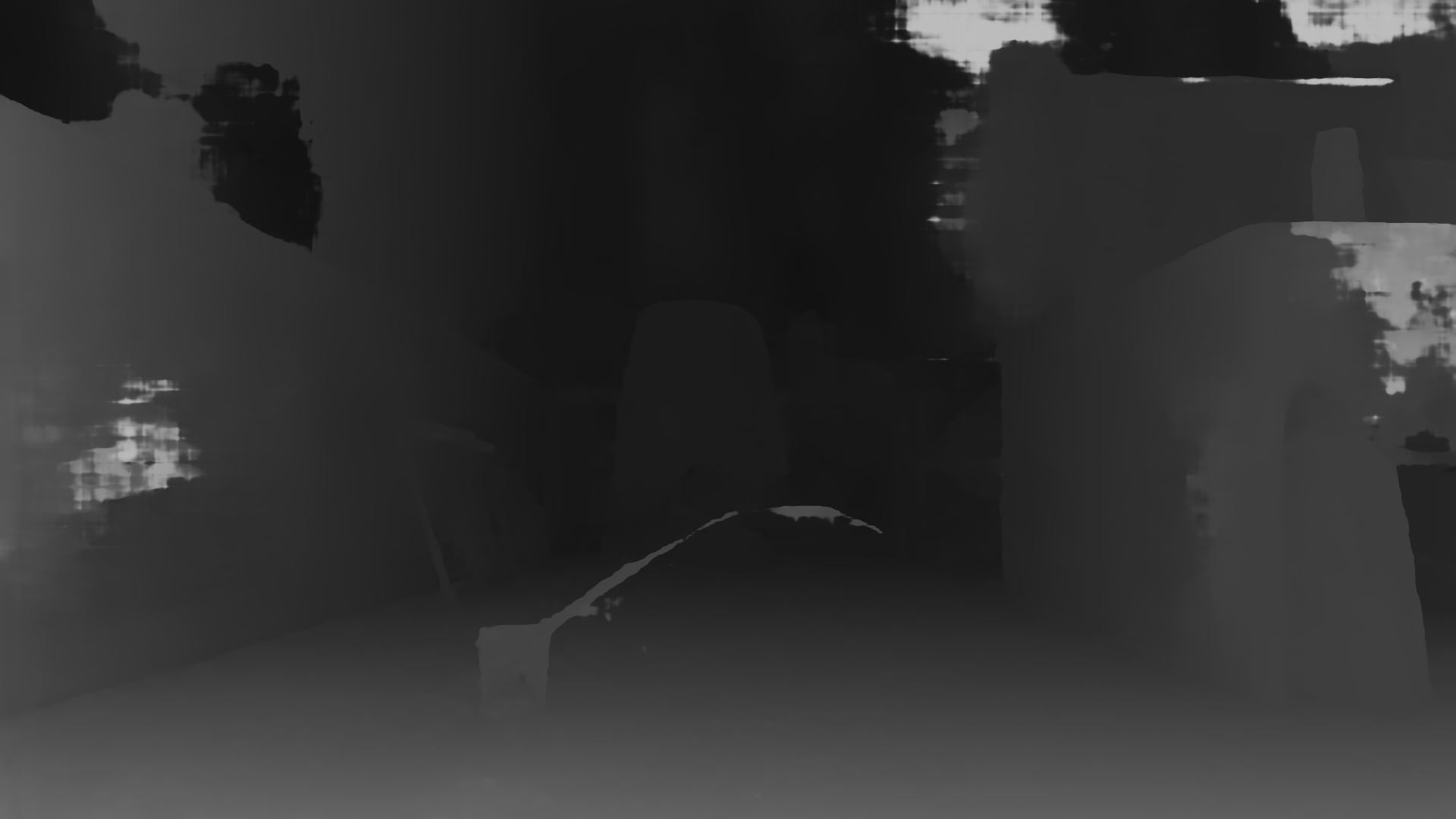}} \subfigure[ACVNet]{\includegraphics[width=0.3\textwidth]{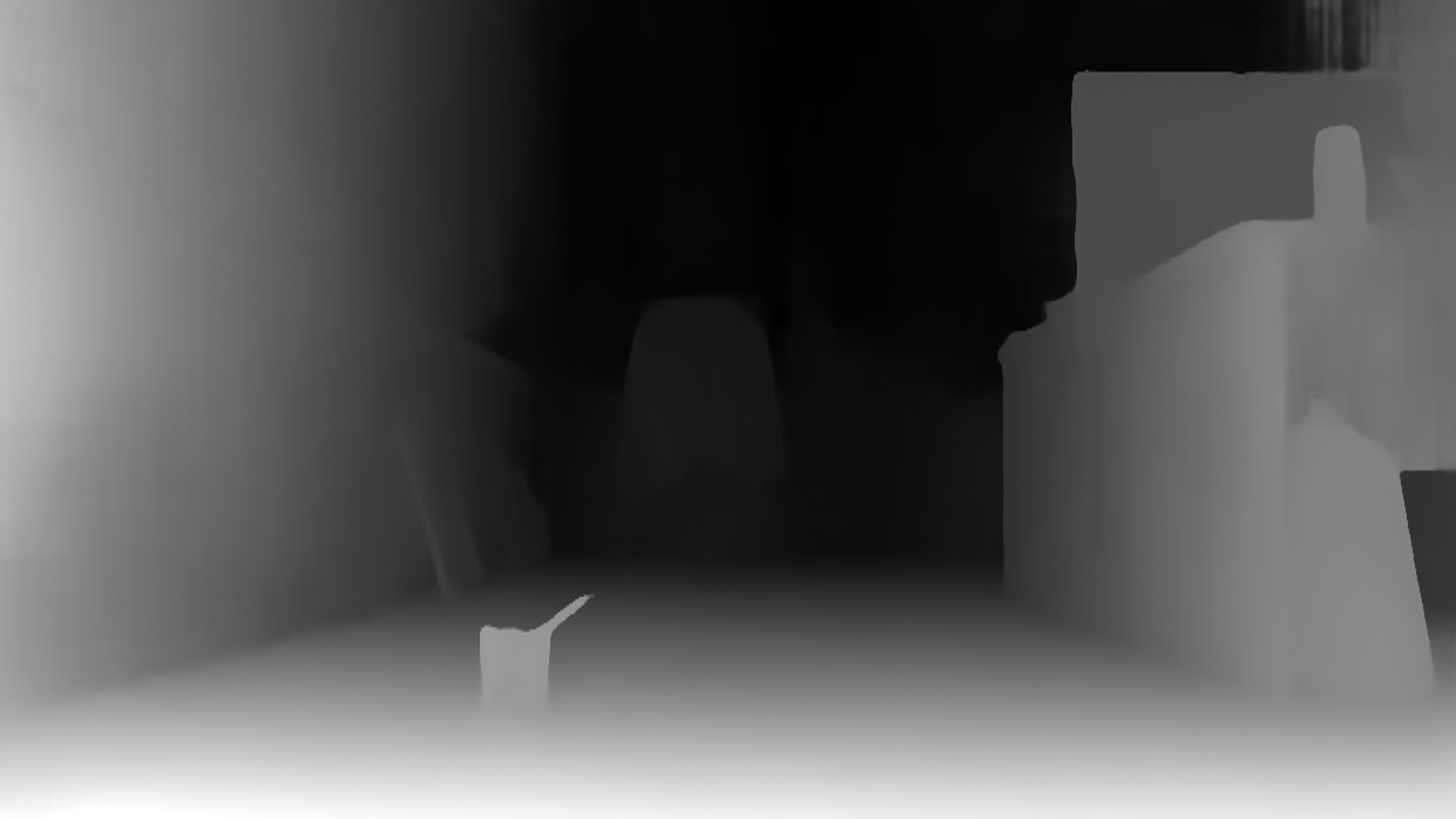}}
        \subfigure[GWCNet]{\includegraphics[width=0.3\textwidth]{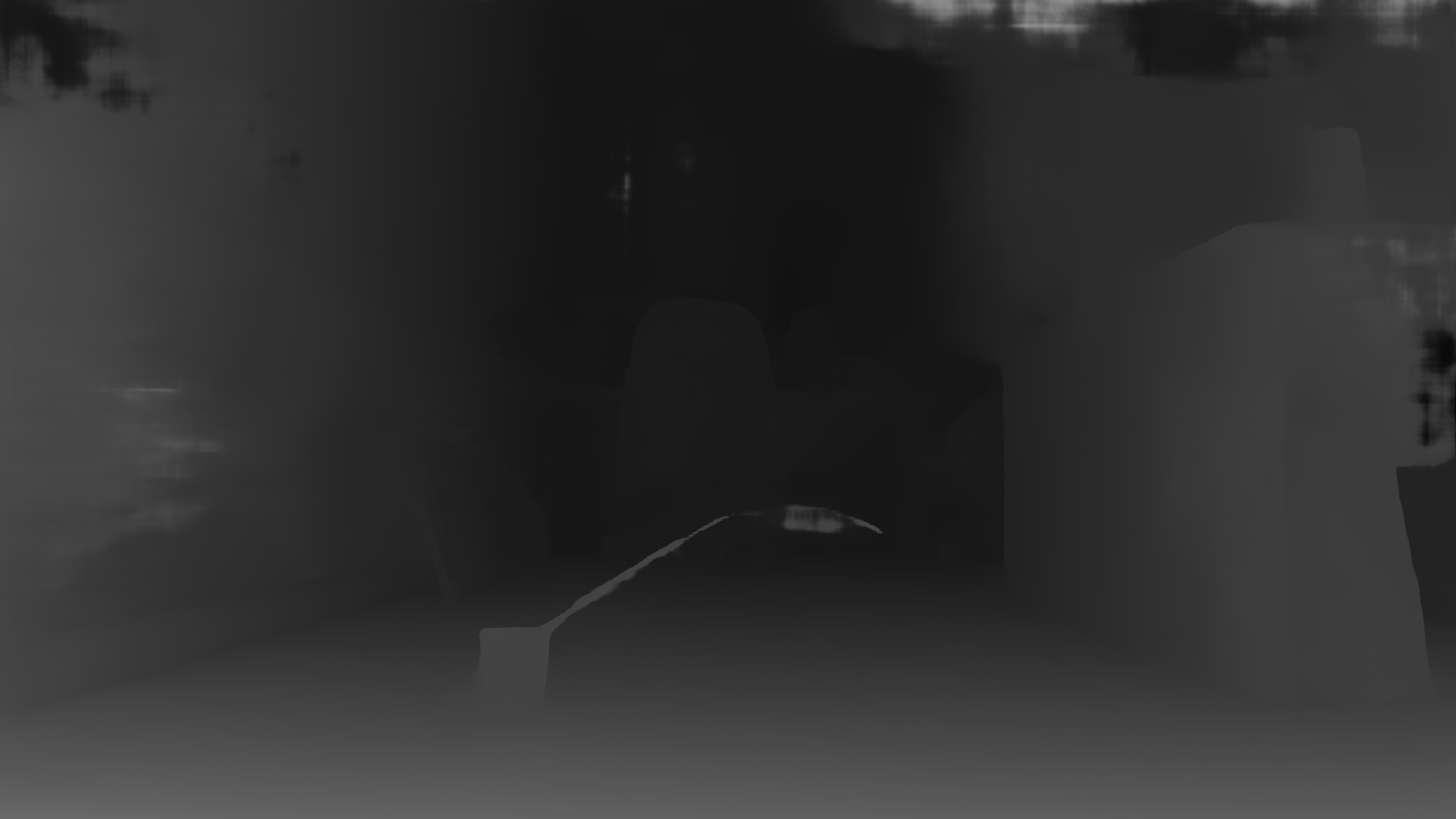}} \subfigure[MobileStereoNet]{\includegraphics[width=0.3\textwidth]{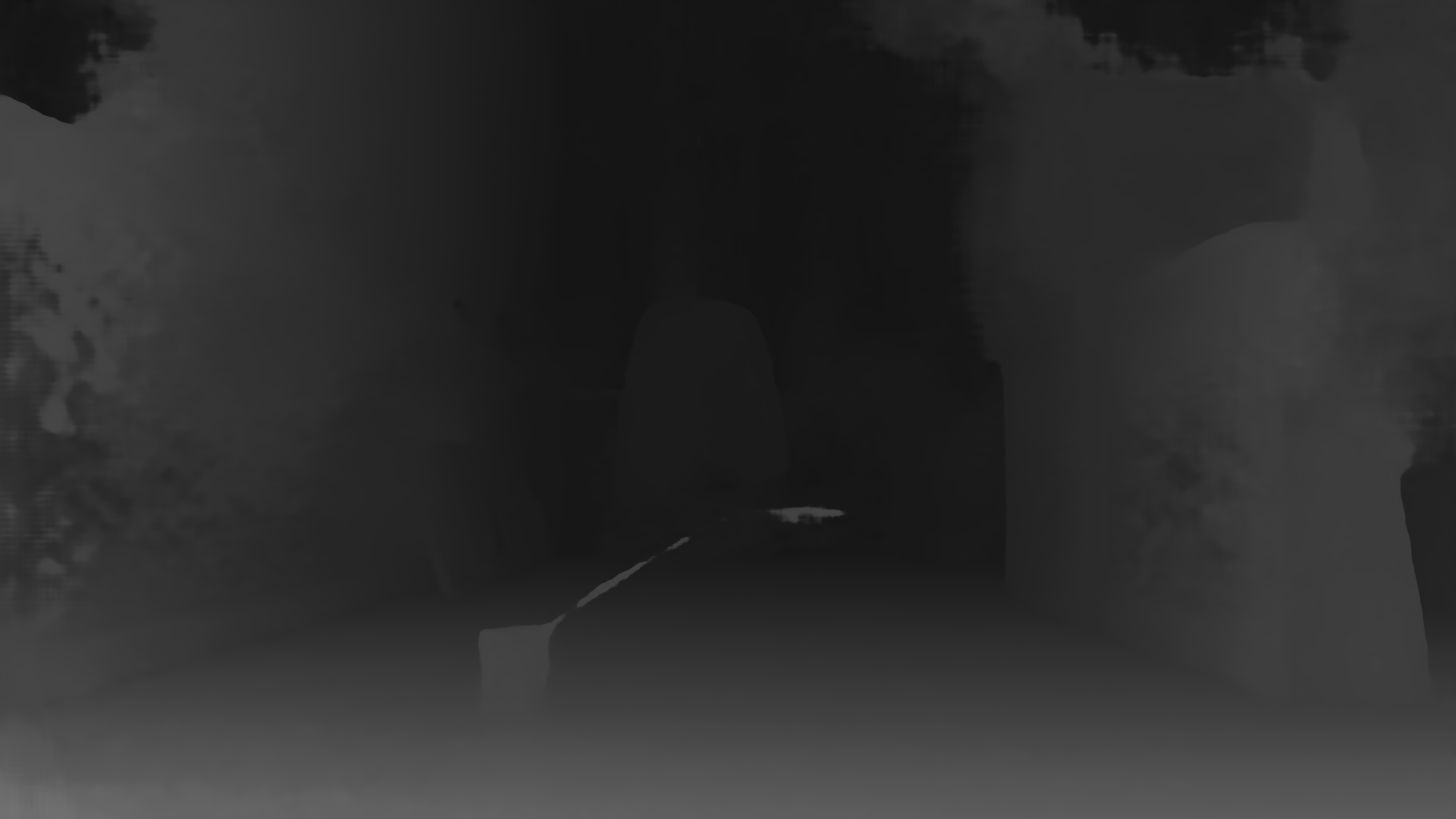}}
        \subfigure[RAFT-Stereo]{\includegraphics[width=0.3\textwidth]{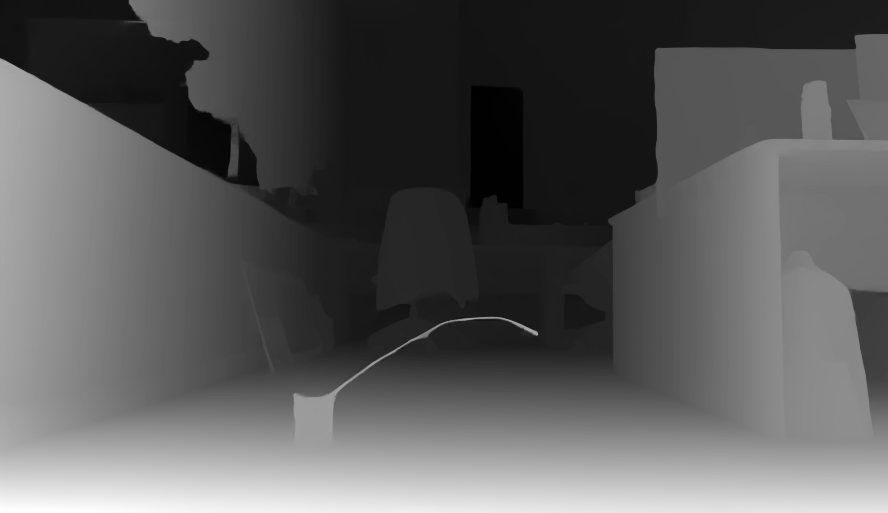}}
        \subfigure[NeRF-Stereo]{\includegraphics[width=0.3\textwidth]{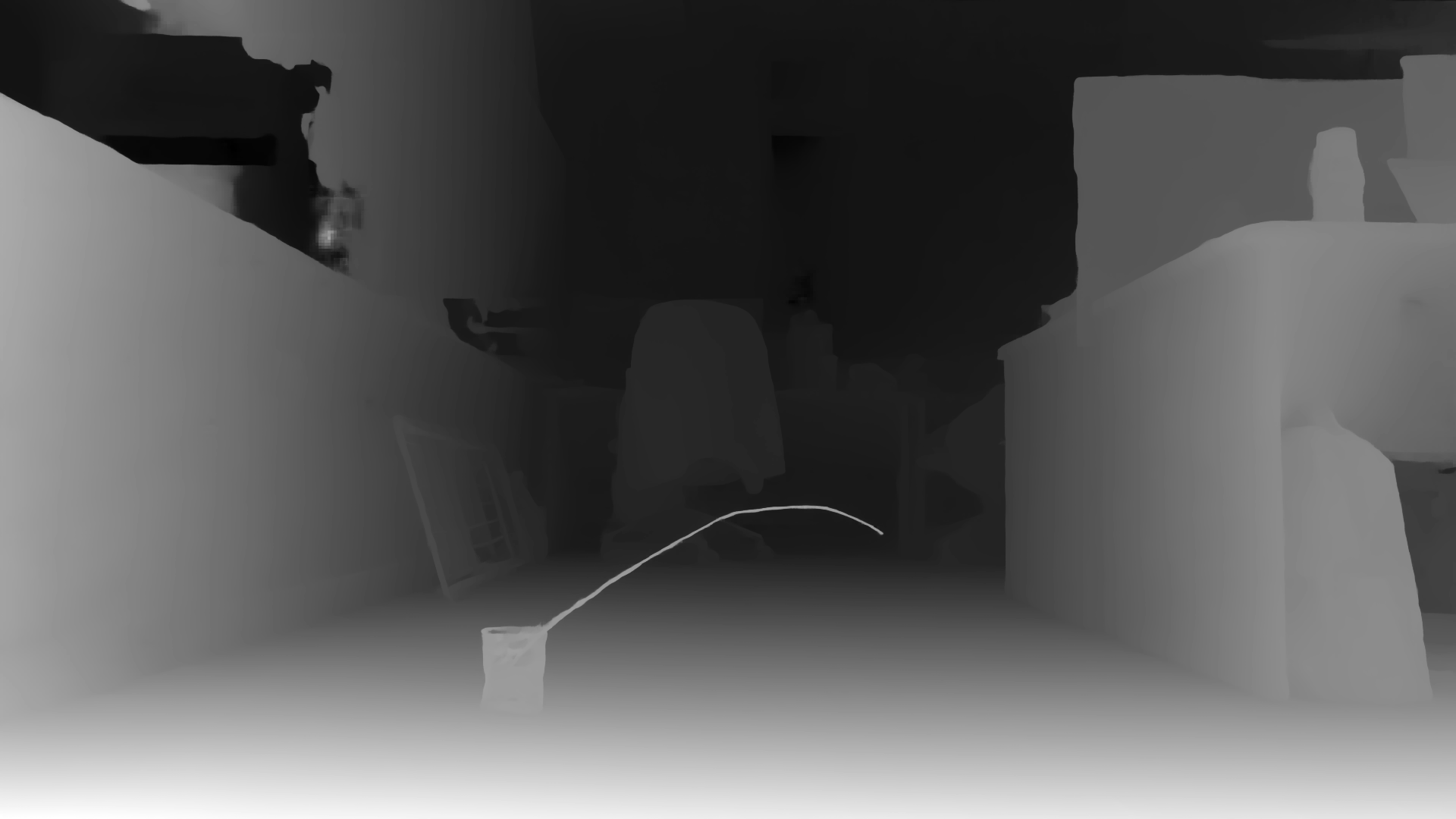}}
        \caption{Disparity maps from six deep stereo architectures on a common
        branch stereo pair.}
        \label{fig:stereo_comparison}
    \end{figure}

    Figure~\ref{fig:stereo_comparison} compares the disparity maps produced by
    all six deep stereo networks on the same branch input.

    Cost-volume networks (PSMNet, GWCNet, ACVNet) recover the global structure of
    the scene but lose detail along the branch silhouette, consistent with the
    resolution loss inherent to 3D-cost-volume aggregation. MobileStereoNet shows
    the same global behaviour with additional smoothing, as expected from its
    compute-constrained design. RAFT-Stereo's iterative refinement preserves the
    branch edges much more crisply, at the cost of a heavier per-frame
    computation. NeRF-Supervised Deep Stereo produces the cleanest branch disparity
    overall and---crucially for this application---does so without requiring in-domain
    ground-truth disparity, which is the main reason it is selected for the
    integrated pipeline below.

    \subsection{End-to-End Branch Distance Estimation}

\begin{figure}[htbp]
        \centering
        \subfigure[detected points on the depth map]{\includegraphics[width=0.45\textwidth]{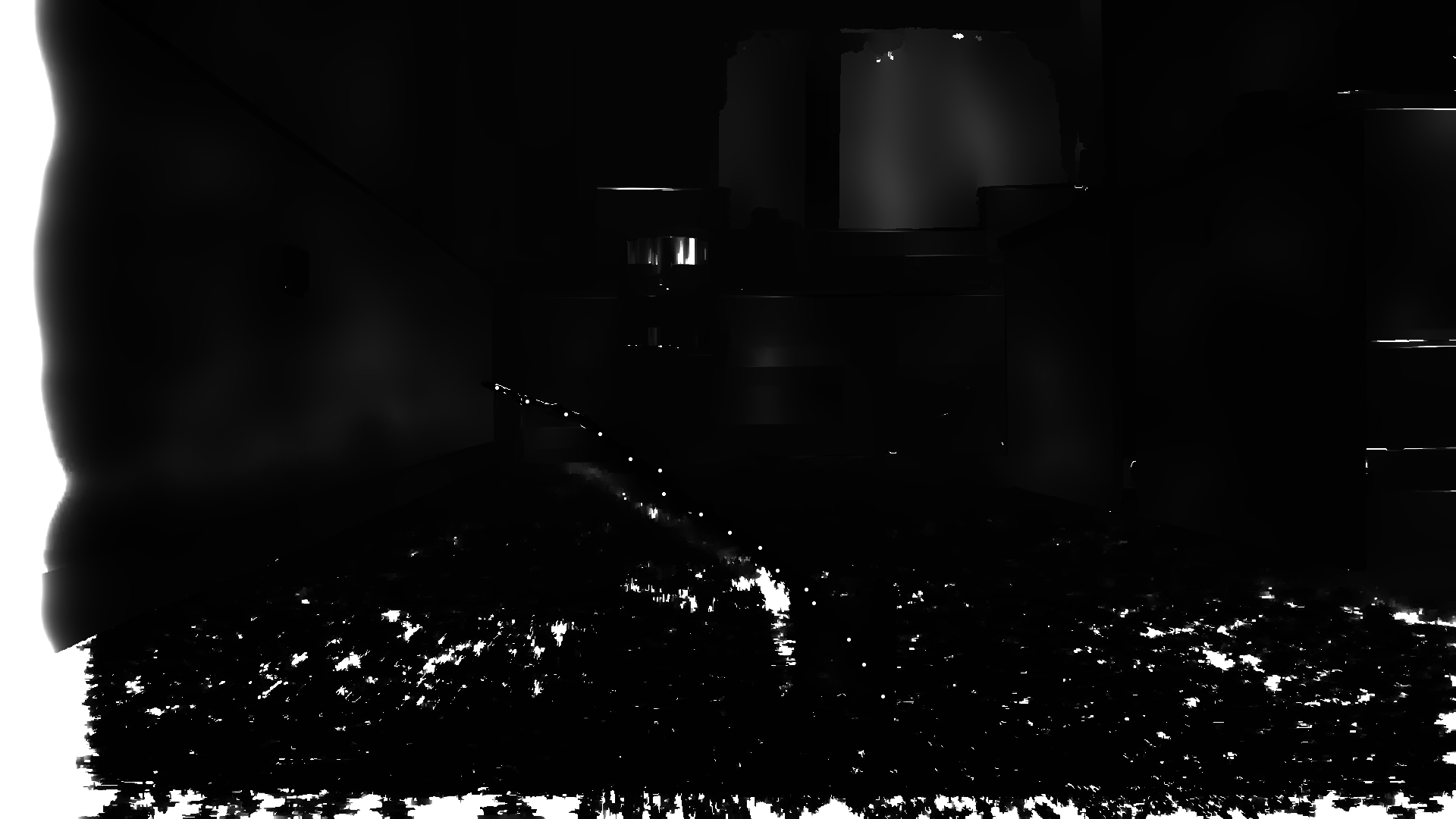}}
        \hfill \subfigure[detected points on the RGB image]{\includegraphics[width=0.45\textwidth]{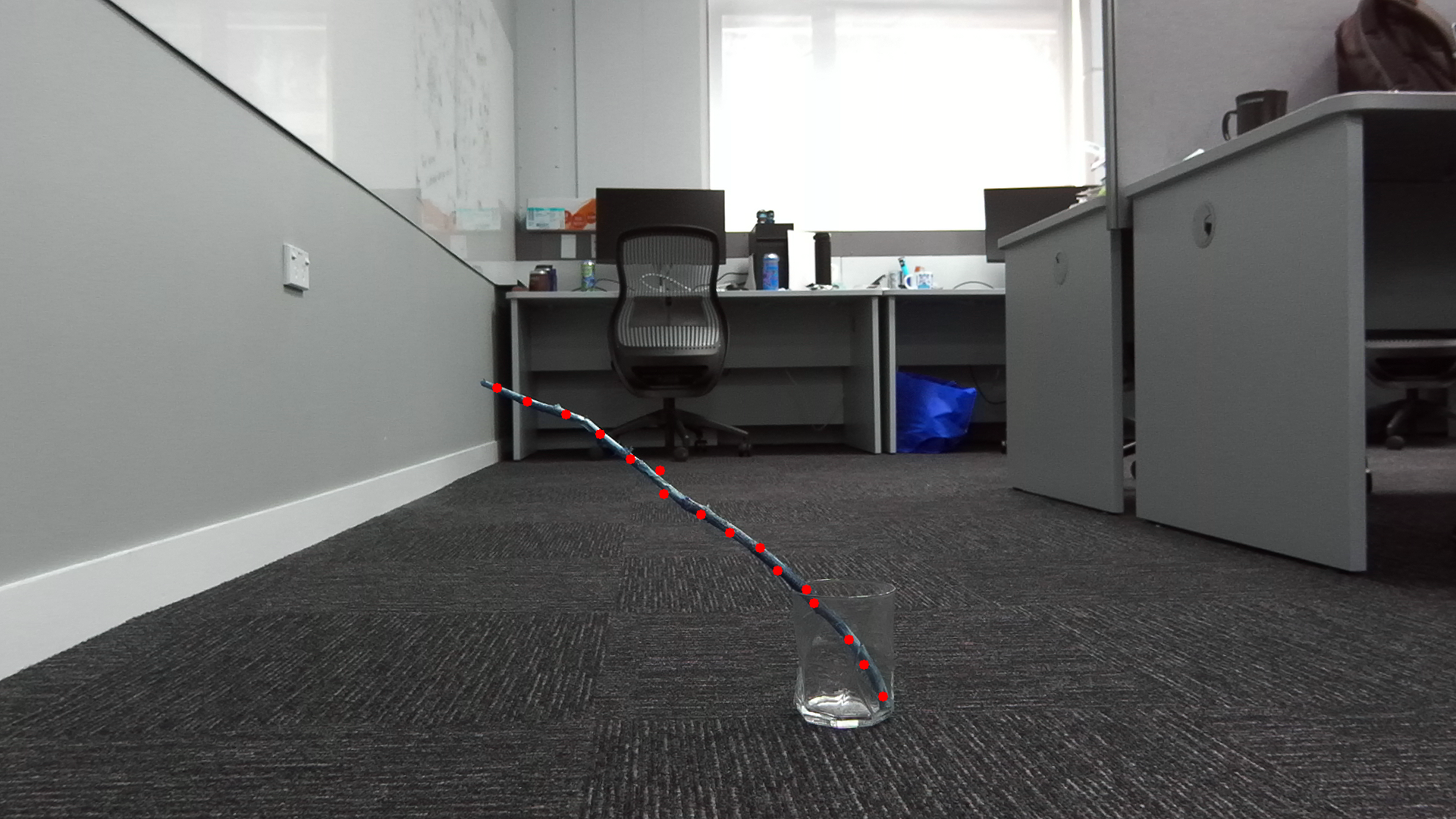}}
        \caption{Branch depth detection results: (a) detected points overlaid on
        the depth map after centroid sampling and MAD filtering, and (b) the
        corresponding points on the RGB image.}
        \label{fig:centroid_depth}
    \end{figure}

    Figure~\ref{fig:centroid_depth} shows the output of the centroid-based triangulation
    with MAD filtering on a representative branch image. Red dots in Figure~\ref{fig:centroid_depth}(a)
    are the locations in $\bar{P}$ that survive MAD filtering on the depth map;
    the same locations overlaid on the RGB image (b) confirm that the retained
    samples lie on the branch interior rather than on the silhouette.

\begin{figure}[htbp]
        \centering
        \subfigure[SGBM depth 1m]{\includegraphics[width=0.3\textwidth]{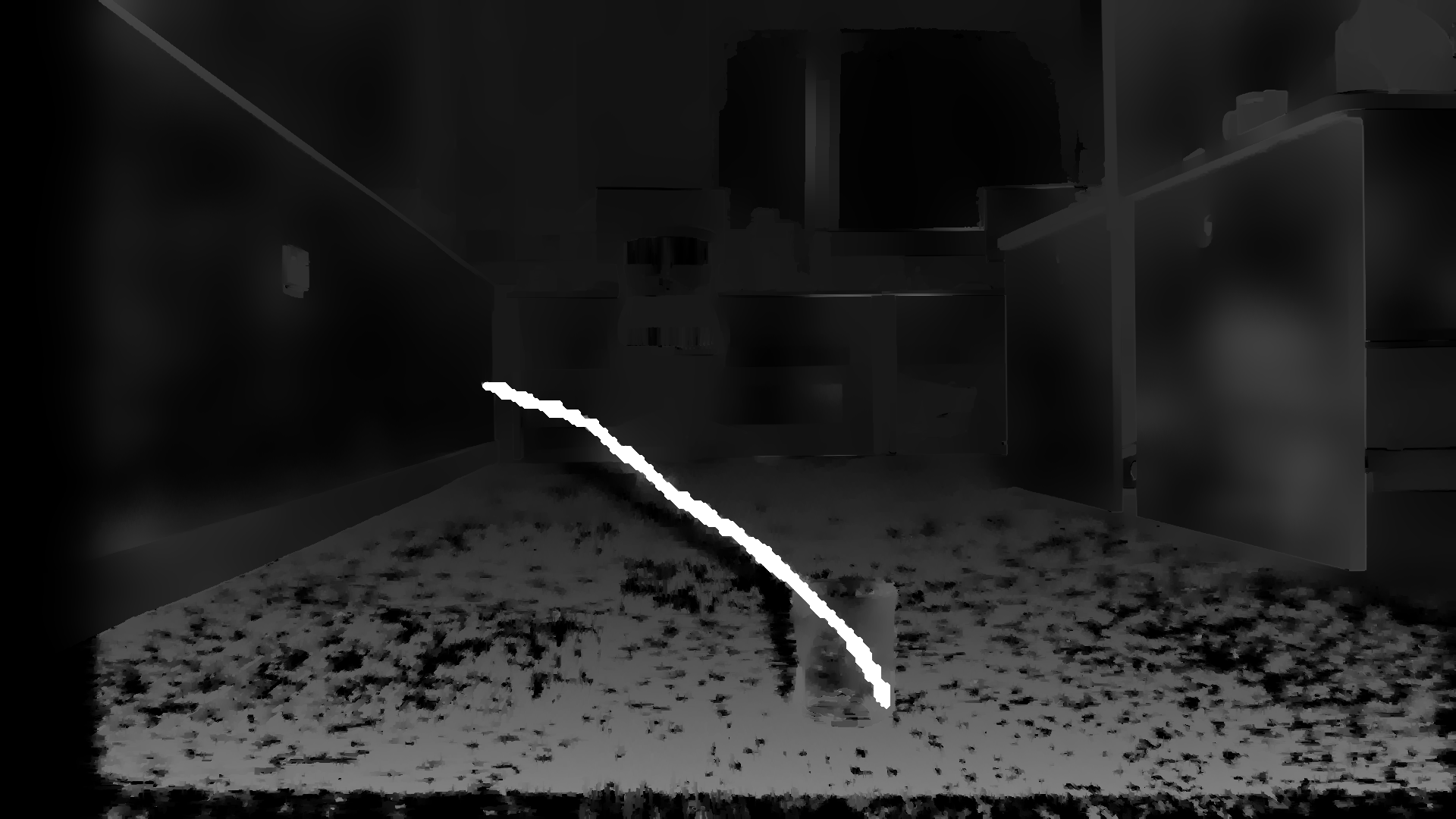}}
        \subfigure[SGBM depth 1.5m]{\includegraphics[width=0.3\textwidth]{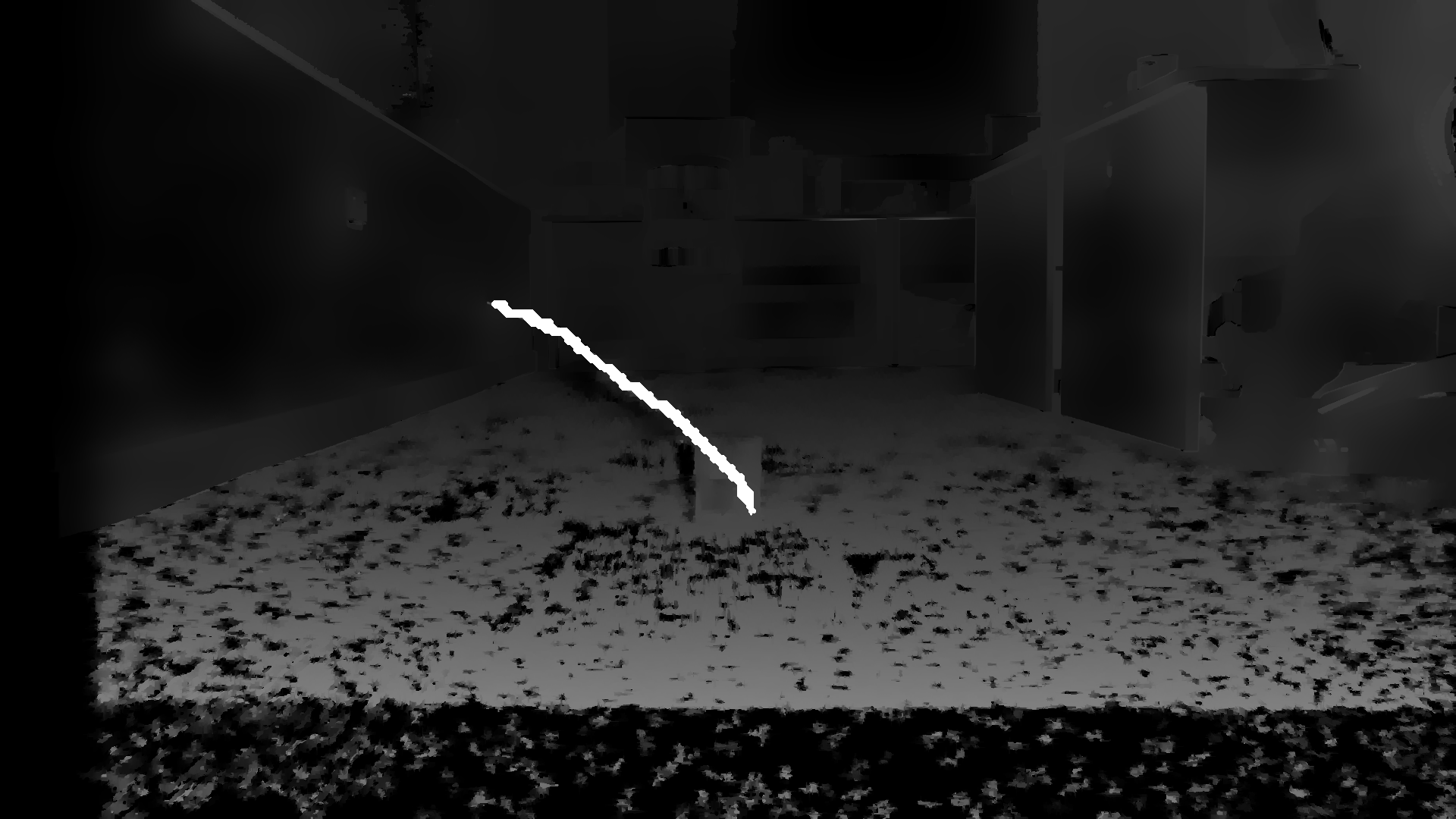}}
        \subfigure[SGBM depth 2m]{\includegraphics[width=0.3\textwidth]{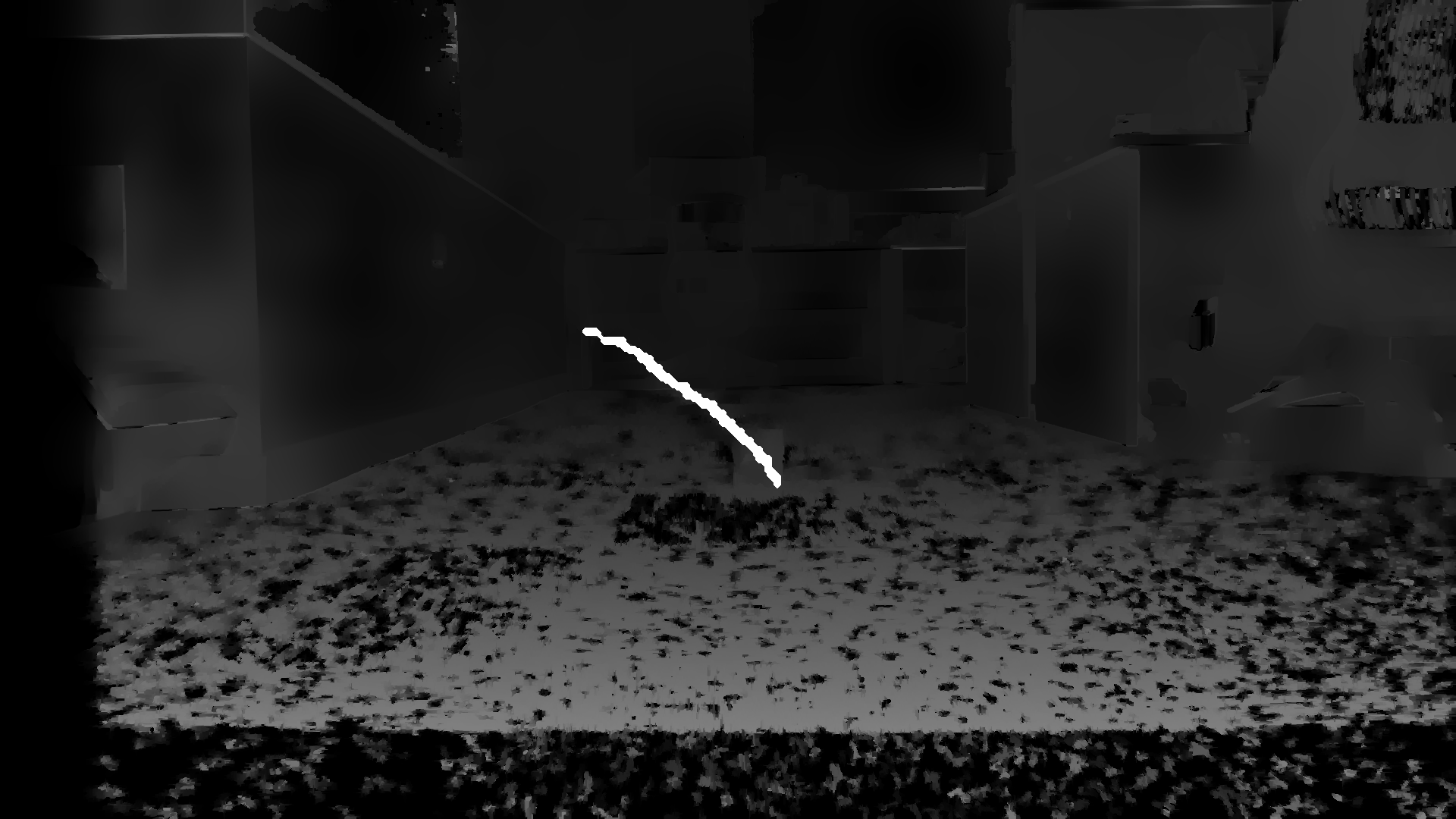}}
        \subfigure[NeRF-Stereo depth 1m]{\includegraphics[width=0.3\textwidth]{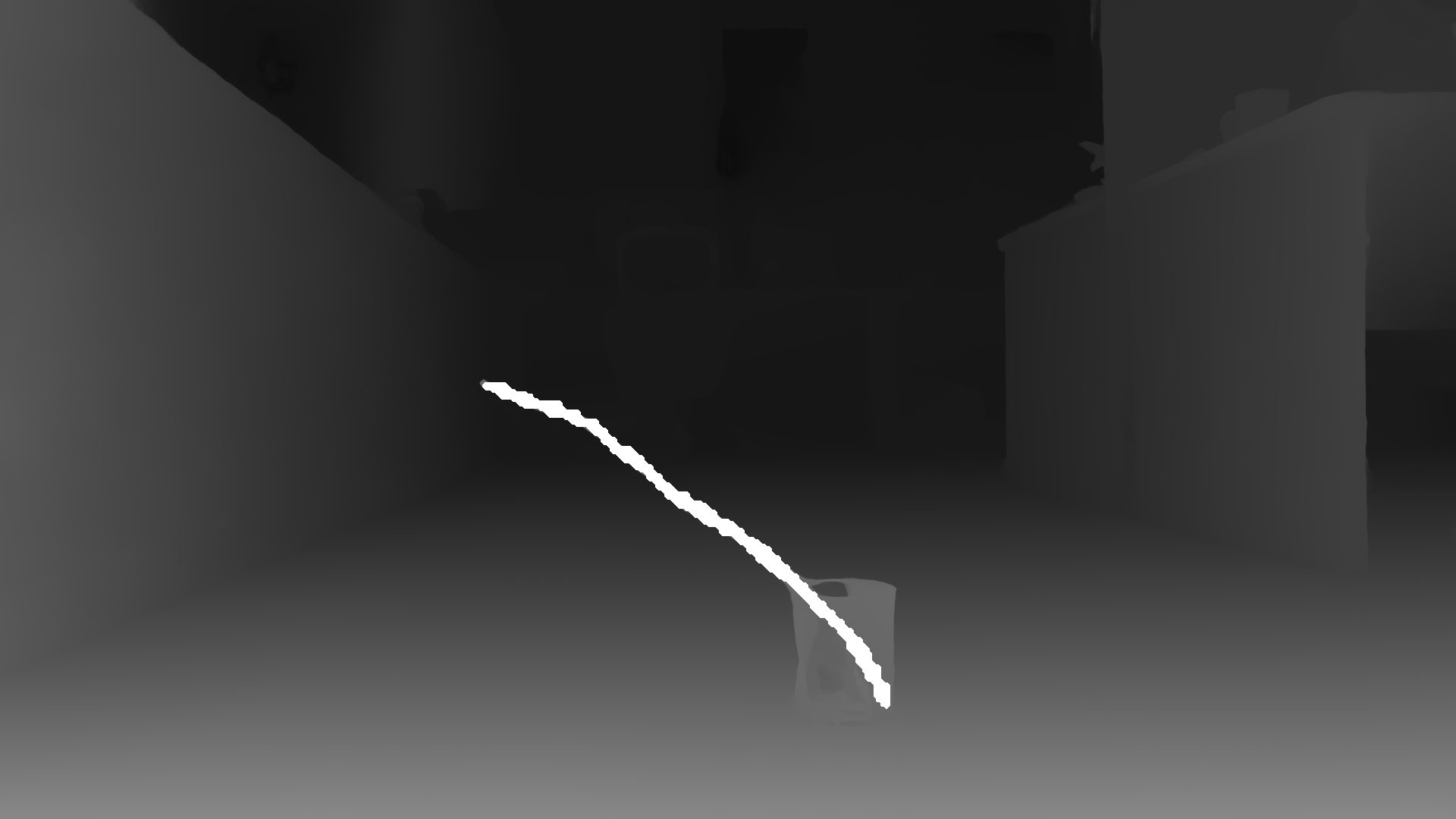}}
        \subfigure[NeRF-Stereo depth 1.5m]{\includegraphics[width=0.3\textwidth]{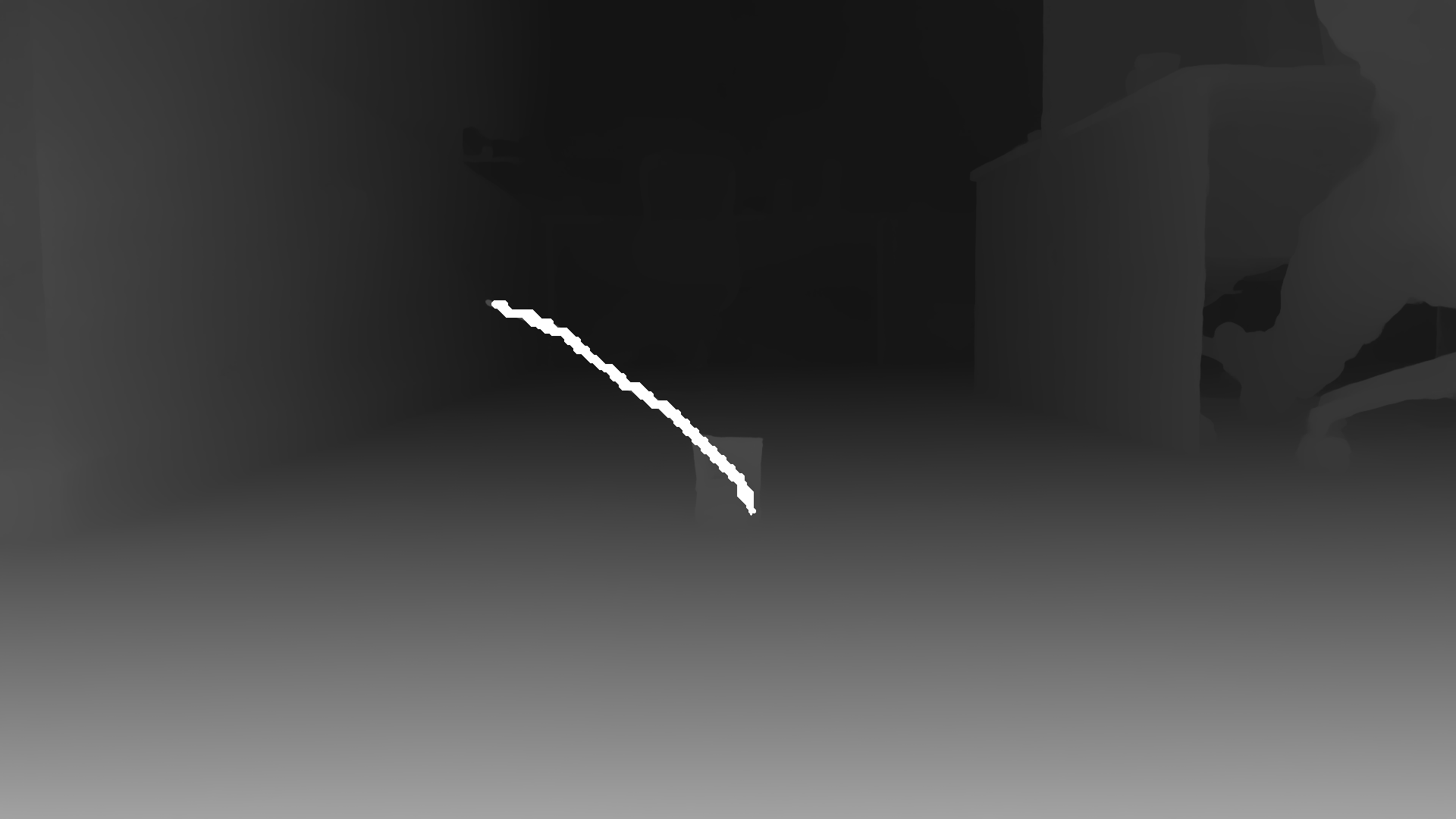}}
        \subfigure[NeRF-Stereo depth 2m]{\includegraphics[width=0.3\textwidth]{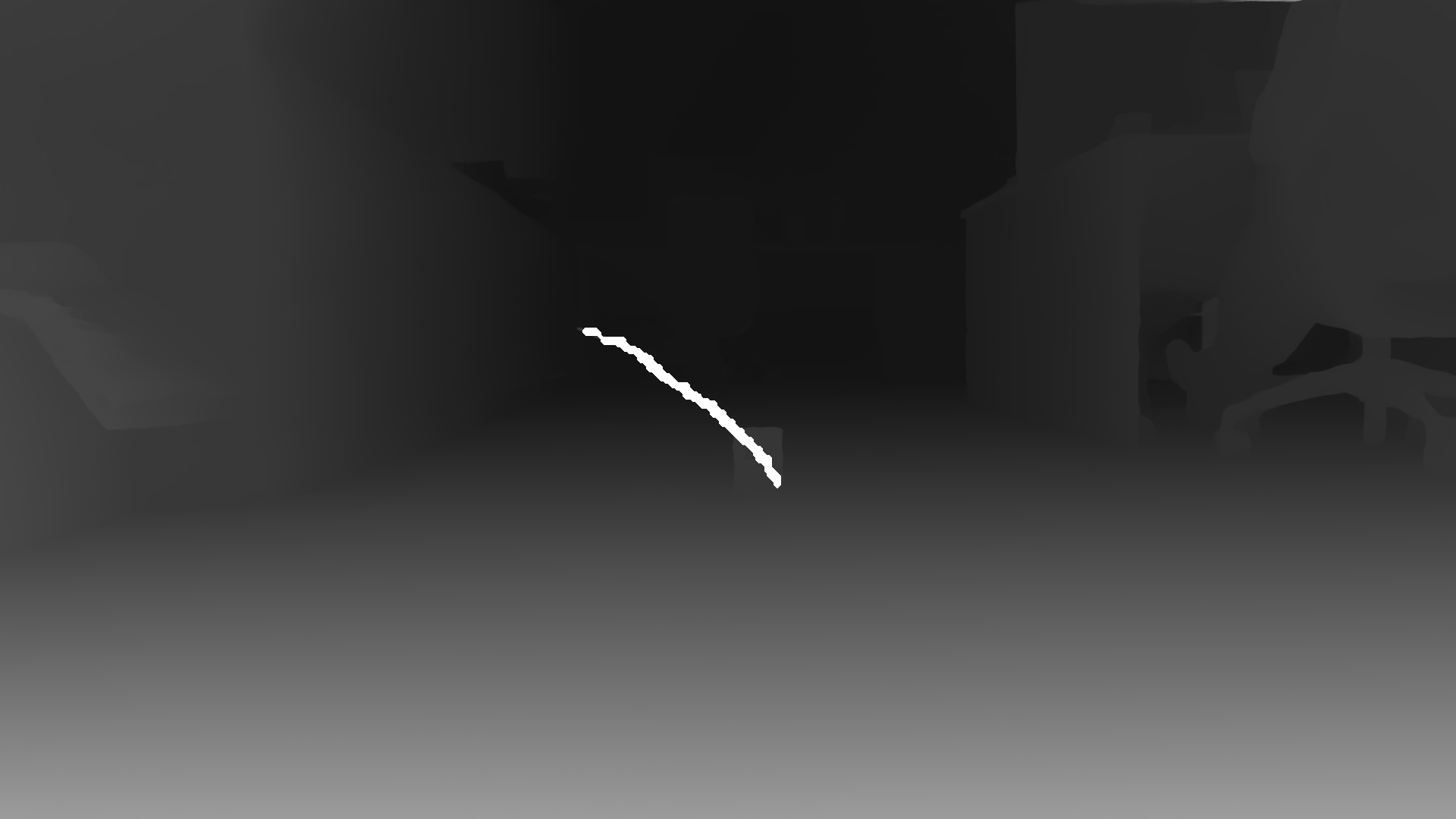}}
        \caption{Depth estimation around the branch using SGBM (top row) and
        NeRF-Stereo (bottom row) at branch--camera distances of 1\,m, 1.5\,m, and
        2\,m.}
        \label{fig:depth_comparison}
    \end{figure}

\begin{figure}[htbp]
        \centering
        \subfigure[SGBM histogram 1m]{\includegraphics[width=0.3\textwidth]{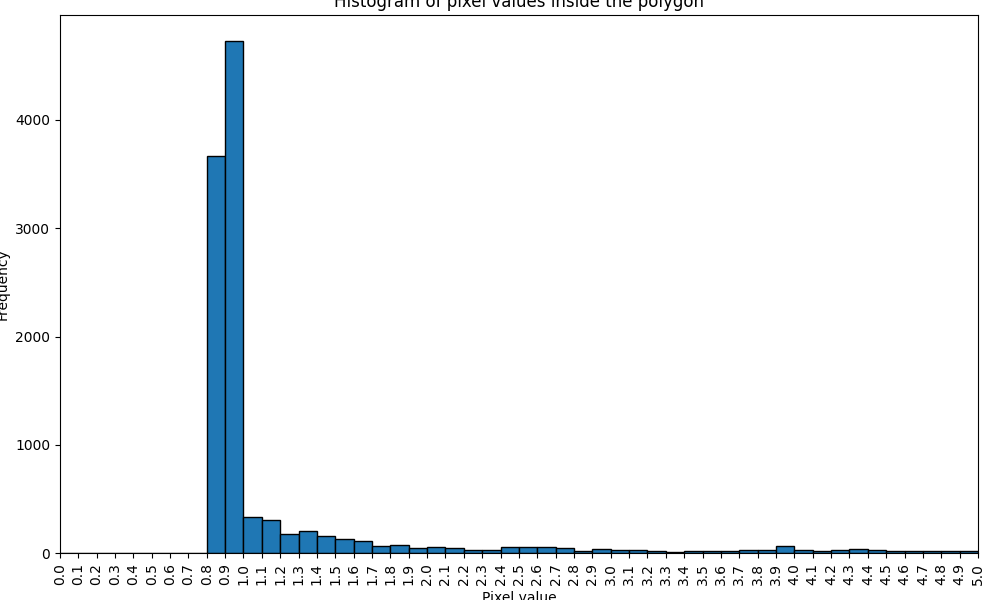}}
        \subfigure[SGBM histogram 1.5m]{\includegraphics[width=0.3\textwidth]{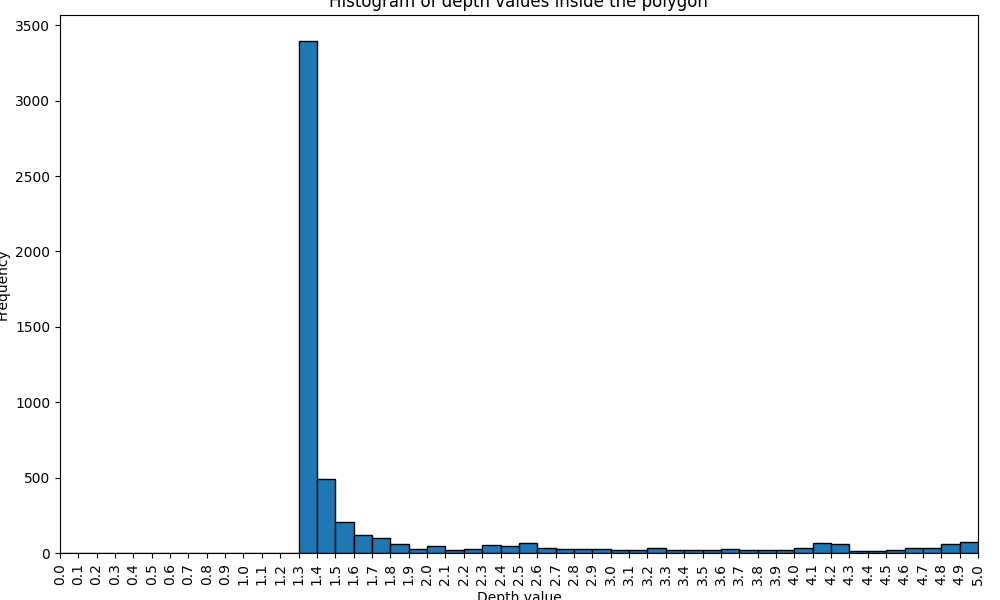}}
        \subfigure[SGBM histogram 2m]{\includegraphics[width=0.3\textwidth]{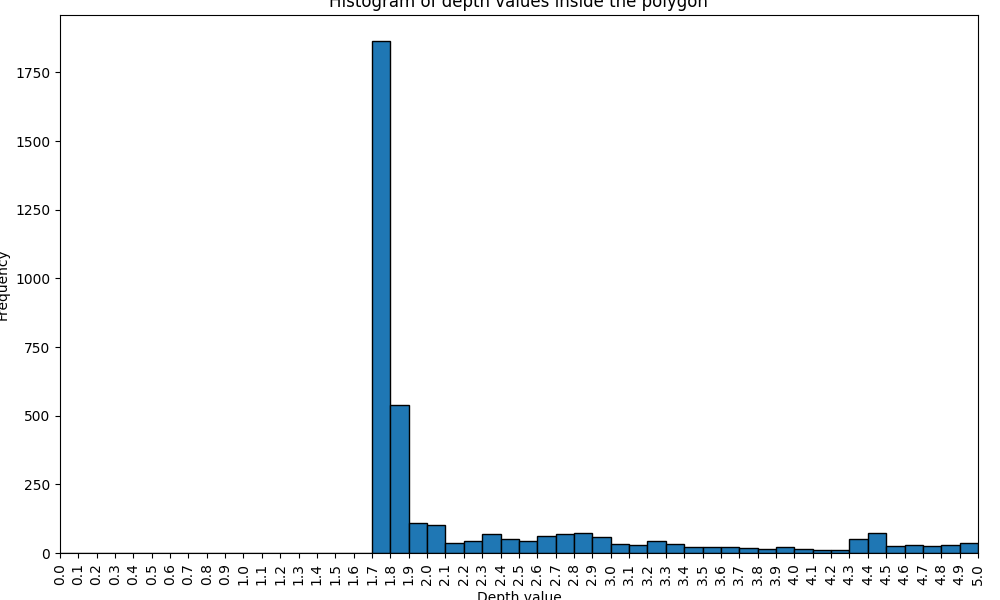}}
        \subfigure[NeRF-Stereo histogram 1m]{\includegraphics[width=0.3\textwidth]{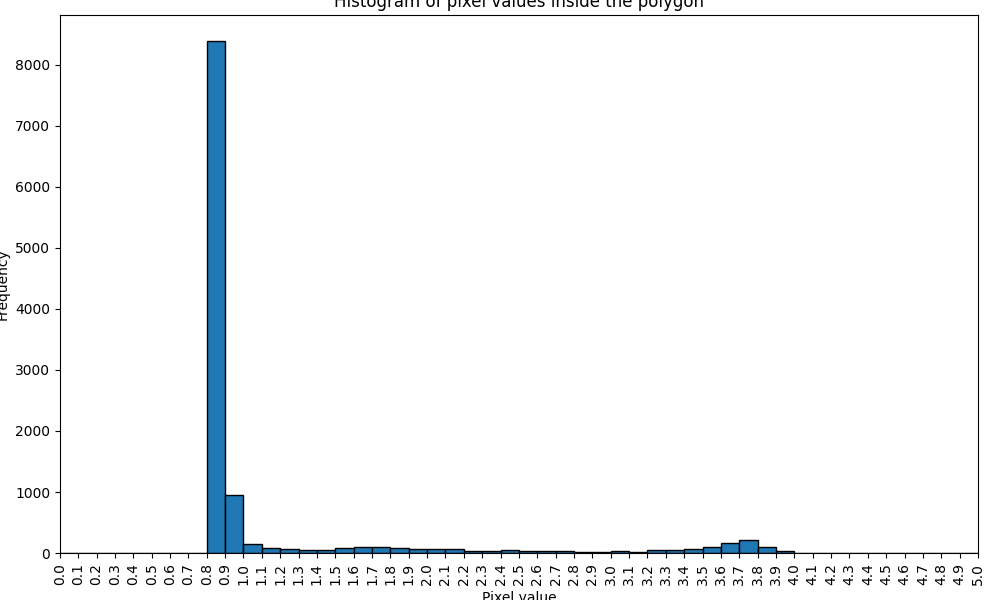}}
        \subfigure[NeRF-Stereo histogram 1.5m]{\includegraphics[width=0.3\textwidth]{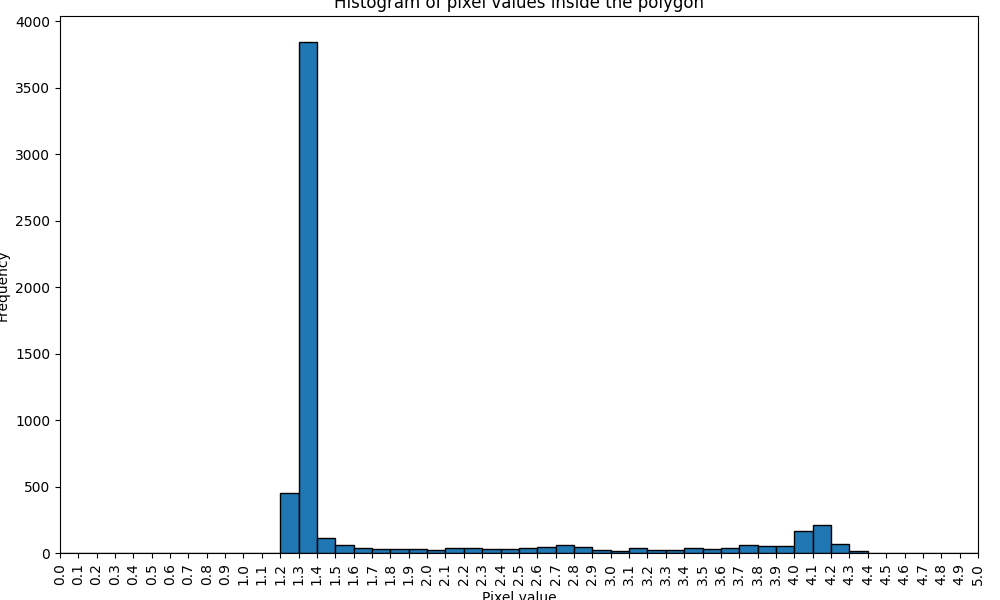}}
        \subfigure[NeRF-Stereo histogram 2m]{\includegraphics[width=0.3\textwidth]{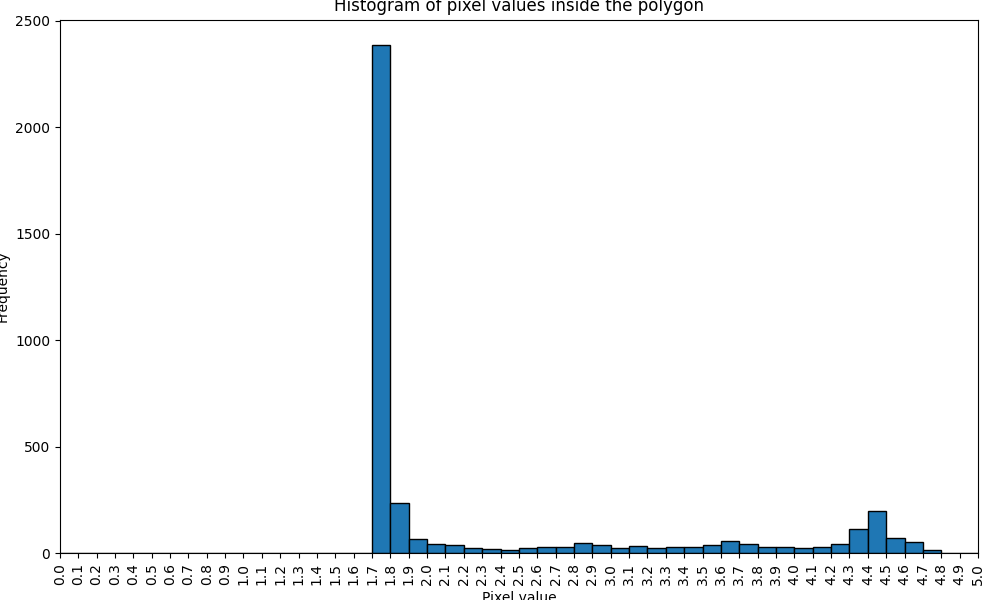}}
        \caption{Histograms of branch-pixel depth values for SGBM (top row) and
        NeRF-Stereo (bottom row) at 1\,m, 1.5\,m, and 2\,m.}
        \label{fig:histograms}
    \end{figure}

    To test hypothesis~H3, the integrated pipeline is run with two different
    stereo back-ends (SGBM with WLS and NeRF-Stereo) at three branch--camera distances
    (1\,m, 1.5\,m, 2\,m). The resulting depth maps are shown in Figure~\ref{fig:depth_comparison},
    and the histograms of per-pixel branch depth---which serve as a proxy for the
    spread of the distance estimate---are shown in Figure~\ref{fig:histograms}.

    Two qualitative observations support H3. First, the NeRF-Stereo histograms at
    all three distances are visibly more concentrated around the true distance than
    the SGBM histograms, which is consistent with the cleaner disparity maps in
    Figure~\ref{fig:stereo_comparison}. Second, both back-ends produce a \emph{single}
    stable distance after centroid sampling with MAD filtering, even though the
    underlying SGBM disparity is visibly noisier; this is the expected effect of
    the MAD-based outlier rejection.

    \subsection{Ablation and Discussion}
    \label{sec:ablation}

    The fusion stage admits three natural ablations whose qualitative behaviour
    can be read directly from the figures above:
    \begin{itemize}
        \item \textbf{No MAD filtering} (mean of all mask pixels). On SGBM, this
            is dominated by the streaking artefacts visible in Figure~\ref{fig:sgbm_pipeline}(e),
            and the resulting distance estimate drifts by tens of centimetres at
            2\,m. On NeRF-Stereo the effect is smaller but still visible as a long
            tail in the histograms of Figure~\ref{fig:histograms}.

        \item \textbf{All-mask sampling instead of centroid sampling.} Sampling
            every mask pixel rather than only centroid neighbourhoods increases
            the influence of silhouette pixels, which tend to bleed into the background.
            The retained-sample overlay in Figure~\ref{fig:centroid_depth} shows
            that the centroid variant deliberately concentrates samples on the
            medial line of the branch and away from the silhouette.

        \item \textbf{SGBM versus deep-stereo back-end.} Replacing SGBM with WLS
            by NeRF-Stereo, with the rest of the pipeline held fixed, produces
            narrower depth histograms at every tested distance (Figure~\ref{fig:histograms}).
            This isolates the contribution of the disparity estimator from the contribution
            of the fusion stage.
    \end{itemize}

    Across all three ablations, the qualitative trend is consistent: the
    centroid-based triangulation with MAD filtering reduces the sensitivity of
    the final distance to individual disparity errors, and a learned stereo back-end
    further tightens the resulting estimate. A formal quantitative ablation, using
    a LiDAR-based reference distance, is currently limited by the indoor-only
    nature of the dataset and is identified as the principal direction for
    future work in Section~\ref{sec:future}.

    \section{Future Work}
    \label{sec:future}

    Three concrete extensions follow directly from the limitations identified above.
    First, the dataset must be extended from the current 71 indoor stereo pairs to
    operational forestry conditions. Outdoor capture from the drone platform
    itself is required to expose the segmentor and the stereo estimator to wind-induced
    branch motion, variable natural illumination, and dense canopy clutter, none
    of which are present in the current indoor set. Where it is possible to
    integrate a LiDAR sensor on the platform, LiDAR returns will be used as a metric
    reference distance so that the qualitative analysis of Section~\ref{sec:results}
    can be replaced by a quantitative end-to-end accuracy figure for the centroid-based
    triangulation with MAD filtering distance estimate, including a formal
    ablation against unfiltered mean-of-mask and against all-mask sampling.

    Second, the deep-stereo back-end will be retrained with in-domain supervision.
    The cross-dataset experiment in Section~\ref{sec:results} shows that KITTI-fine-tuned
    PSMNet does not transfer to thin branches; the natural alternative is either
    Scene~Flow pre-training followed by self-supervised adaptation on unlabelled
    forestry stereo pairs, or NeRF-Supervised Deep Stereo trained directly on
    radiance fields built from the captured pine sequences. Both paths avoid the
    cost of dense disparity annotation in the field.

    Third, the present pipeline runs segmentation and stereo as two independent
    networks. A multi-task architecture that shares an encoder and emits both an
    instance mask and a disparity map in a single forward pass would reduce on-board
    latency and is the most promising route to real-time operation on a drone payload.
    Newer YOLO releases that appeared after the data-collection cut-off can be
    substituted into the segmentation branch without changes to the fusion stage.

    \section{Conclusion}
    \label{sec:conclusion}

    This paper has investigated whether a single low-cost stereo camera can deliver
    branch detection and 3D positioning of sufficient quality to support autonomous
    pruning of radiata pine branches as thin as 10\,mm at working distances of 1--2\,m,
    without auxiliary depth sensors. Three findings stand out. (i) On the 10-pair
    branch test set, YOLOv8 and YOLOv9 segmentation variants reach 98.7--99.6 mAP\textsubscript{box50--95}
    and 77.1--82.0 mAP\textsubscript{mask50--95}, while seven Mask~R-CNN backbones
    collapse to below 12 mAP\textsubscript{mask50--95} despite competitive box accuracy;
    thin, elongated branches are therefore much better served by prototype-based
    one-stage segmentors than by the low-resolution mask head of Mask~R-CNN. (ii)
    Cross-dataset fine-tuning of PSMNet on KITTI~2012 and KITTI~2015 degrades, rather
    than improves, branch disparity, because the urban driving regime differs sharply
    from the thin near-vertical structures of forestry; among the six deep
    stereo networks evaluated, NeRF-Supervised Deep Stereo produces the cleanest
    branch disparity without requiring in-domain ground truth. (iii) The
    proposed centroid-based triangulation with MAD outlier rejection turns a
    noisy disparity map into a single stable branch-to-camera distance: depth histograms
    at 1\,m, 1.5\,m and 2\,m are visibly more concentrated around the true
    distance for NeRF-Stereo than for SGBM with WLS, and the same fusion stage absorbs
    the streaking artefacts that dominate the raw SGBM output.

    Taken together, these results indicate that low-cost passive stereo, combined
    with a prototype-based segmentor and the proposed fusion stage, is a viable
    perception front-end for drone-based pruning of fine branches. The principal
    remaining gap, addressed in Section~\ref{sec:future}, is a quantitative outdoor
    evaluation against a metric reference such as LiDAR.

    %我們這個是綫性的，所以時間相加，如何讓其平行，同時運行，未來的工作。
    %照片得到，兩個方向，2個模型分開處理，再結合在一起。
    %照片得到一個模型，同時處理2個任務
    %硬件可以處理，那麽有沒有辦法軟件也這樣處理？？？
    %网络的一部分可能用来处理图像的特征提取，而这些特征同时用于预测每个像素的类别（用于分割）和每个像素的深度（用于深度估计）。这使得模型能够在单次推理中同时输出分割图和深度图。

    \bibliographystyle{plainnat}
    \bibliography{interactcsesample}
\end{document}